\documentclass{article}
\PassOptionsToPackage{numbers, sort, compress}{natbib}
\usepackage[main, final]{neurips_2026}

\usepackage[utf8]{inputenc} 
\usepackage[T1]{fontenc}    
\usepackage{hyperref}       
\usepackage{url}            
\usepackage{booktabs}       
\usepackage{amsfonts}       
\usepackage{nicefrac}       
\usepackage{microtype}      
\usepackage{xcolor}         

\usepackage{float}
\usepackage{algorithm}
\usepackage{algpseudocode}
\usepackage{graphicx}
\usepackage{booktabs}
\usepackage{multirow}
\usepackage{makecell}
\usepackage[table]{xcolor}
\usepackage{enumitem}
\usepackage{amsmath}

\makeatletter\renewcommand{\@fnsymbol}[1]{\ifcase#1\or\dagger\else\@ctrerr\fi}\makeatother

\title{Focused Forcing: Content-Aware Per-Frame KV Selection for Efficient Autoregressive Video Diffusion}

\author{
\textbf{Peiliang Cai}$^{1}$\hspace{0.25cm}
\textbf{Evelyn Zhang}$^{1}$\hspace{0.25cm}
\textbf{Jiacheng Liu}$^{1,2}$\hspace{0.25cm}
\textbf{Hao Lin}$^{3}$\hspace{0.25cm}
\textbf{Ruiqi Zhang}$^{4}$\hspace{0.25cm}
\textbf{Weile Mo}$^{1}$\hspace{0.25cm} \\
\textbf{Yue Ma}$^{5}$\hspace{0.25cm}
\textbf{Shikang Zheng}$^{1,6}$\hspace{0.25cm}
\textbf{Jiehang Huang}$^{1}$\hspace{0.25cm}
\textbf{Dongrui Liu}$^{7}$\hspace{0.25cm}
\textbf{Linfeng Zhang}$^{1}$\thanks{Corresponding author.}
\vspace{2mm}\\
$^{1}$SJTU\hspace{0.25cm}
$^{2}$SDU\hspace{0.25cm}
$^{3}$HUST\hspace{0.25cm}
$^{4}$UTokyo\hspace{0.25cm}
$^{5}$HKUST\hspace{0.25cm}
$^{6}$SCUT\hspace{0.25cm}
$^{7}$Shanghai AI Lab
}

\begin{document}
\raggedbottom
\maketitle

\begin{figure}[htbp]
    \vspace{-6mm}
    \centering
    \includegraphics[width=\linewidth]
    {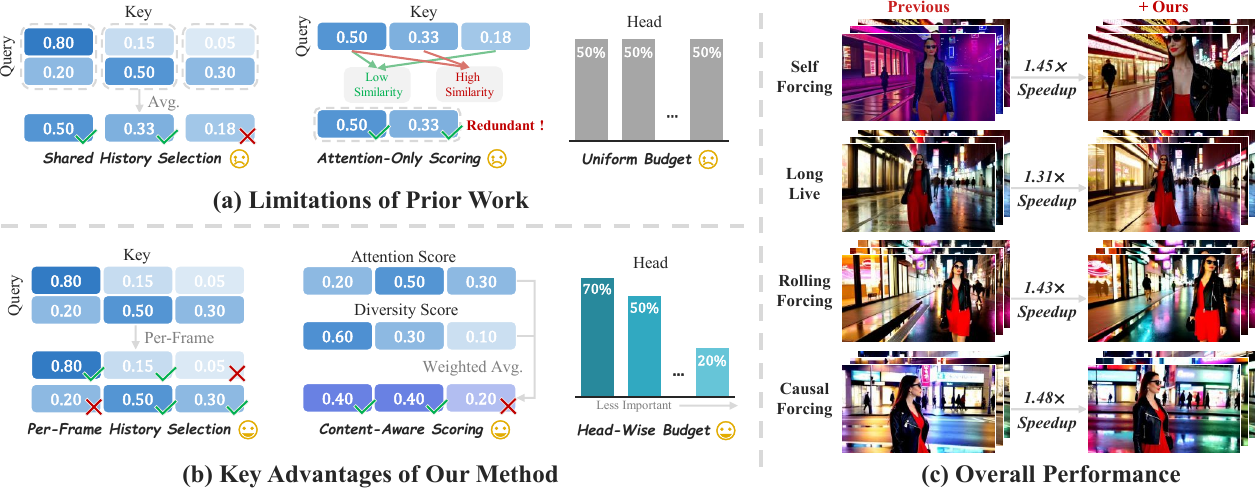}
    \vspace{-6mm}
    \caption{\textbf{Overview of Focused Forcing.} (a) Existing methods use shared history selection, attention-only scoring, and coarse head budgeting, illustrated here with uniform budgets. (b) Focused Forcing uses per-frame history selection, content-aware scoring, and head-adaptive budgets. (c) Focused Forcing achieves up to 1.48$\times$ speedup across multiple paradigms while preserving quality.}
    \label{fig:teaser}
    \vspace{-2mm}
\end{figure}

\begin{abstract}
\vspace{-2mm}
Recent advances in autoregressive video diffusion have enabled sequential and streaming video generation. However, long-horizon generation requires increasingly large KV caches, making efficient compression without sacrificing quality challenging. Existing methods mostly select historical frames based on attention scores, but their context decisions remain coarse. When multiple frames are generated in the same chunk, these methods often apply a shared history selection to the whole chunk, score historical frames solely by attention, and assign head-wise budgets either uniformly or by attention-pattern heuristics rather than explicit head-importance estimation. We show that frames within the same generated chunk can depend on distinct historical frames, that the same historical frame can receive different attention scores as its relative temporal distance to the current frames changes, and that masking different heads induces unequal generation degradation. Motivated by these findings, we propose \textbf{Focused Forcing}, a training-free KV selection method that focuses cached history along both generated-frame and head dimensions. For each generated frame, Focused Forcing preserves the most relevant and distinctive historical frames by combining attention scores with diversity scores of historical frames, while assigning larger budgets to heads with higher estimated importance. Across multiple autoregressive generation paradigms, Focused Forcing achieves up to $\textbf{1.48}\times$ end-to-end acceleration without training, while \textbf{improving visual quality and text alignment}. \textit{Our code is available at} {\texttt{\textcolor{cyan}{https://github.com/caipeiliang2004/Focused-Forcing}}}
\end{abstract}
\section{Introduction}
\label{introduction}

Recent advances in video diffusion models~\cite{wan2_1,ltx-video,cogvideox,ma2024follow, brooks2024video} have greatly improved visual fidelity and motion coherence, making long-form, interactive, and streaming video generation increasingly practical. Unlike bidirectional paradigms that synthesize an entire clip jointly, autoregressive video diffusion~\cite{causvid,diffusion_force} generates video sequentially, often in chunks, and naturally supports online rollout with KV caching~\cite{self_forcing,rolling_forcing,causal_forcing}. However, long-horizon rollout requires maintaining an ever-growing cache of previously generated frames, making later attention operations increasingly expensive in memory and computation.

To improve efficiency, existing methods typically restrict or compress the accessible history with sliding-window or rolling-KV designs~\cite{self_forcing,self_forcing++,context_force,causal_forcing}. Some preserve temporal continuity through structured cache rules such as attention sinks or cache re-caching~\cite{deep_forcing,longlive}, while others select or prune cached frames by attention scores~\cite{deep_forcing}. Recent work also explores head-specific cache strategies~\cite{dummy_forcing}. Despite these advances, as illustrated in Fig.~\ref{fig:teaser}(a), existing methods still make coarse-grained context decisions in three aspects: \textbf{1) Shared Chunk-Level Selection}, where all frames in a generated chunk share the same history selection; \textbf{2) Attention-Only Ranking}, ranking historical frames mainly by attention scores; and \textbf{3) Coarse Head-Wise Budgeting}, allocating head-wise budgets uniformly or heuristically without explicit head-importance estimation. These designs reduce cache overhead, but do not fully match the fine-grained context requirements of autoregressive video generation.

\begin{figure}[htbp]
    \vspace{-2mm}
    \centering
    \includegraphics[width=\linewidth]
    {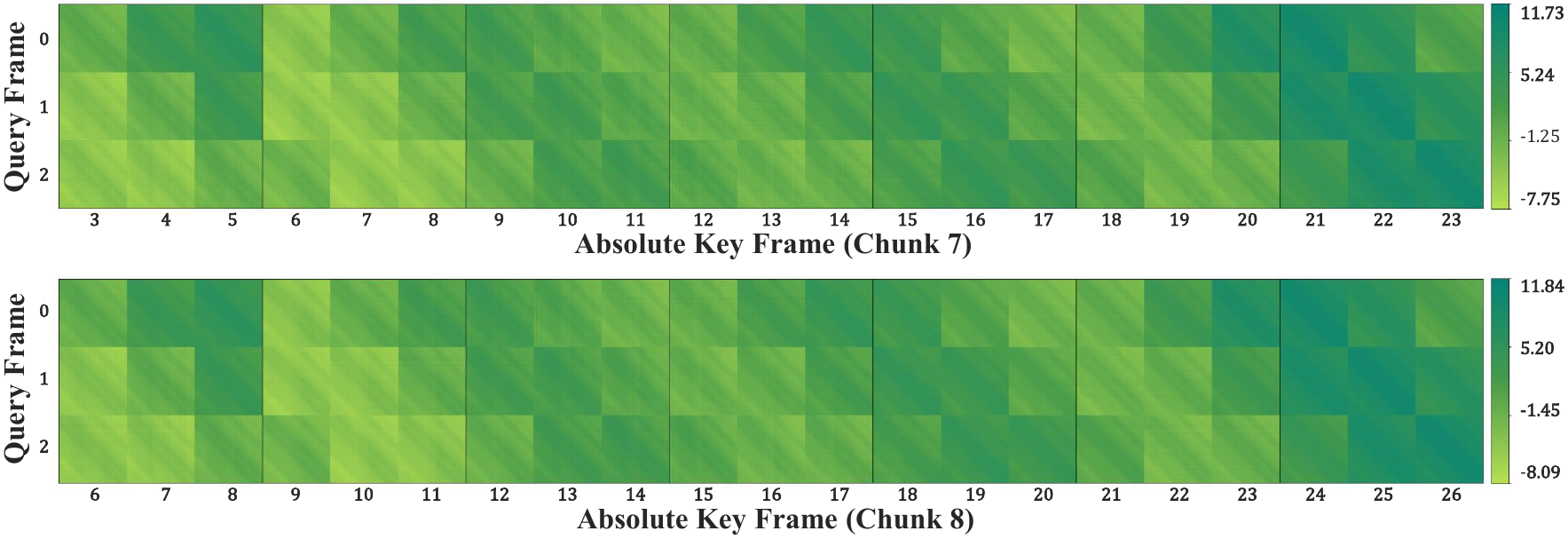}
    \vspace{-6mm}
    \caption{\textbf{Attention is generated-frame-dependent and relative-temporal-distance-sensitive.} Rows denote query frames in the current generated chunk, and columns denote historical key frames. Attention varies across query frames and changes with relative temporal distance.}
    \label{fig:attn_pattern}
    \vspace{-2mm}
\end{figure}

Attention patterns in autoregressive video diffusion suggest that history selection should be performed at a finer granularity. As shown in Fig.~\ref{fig:attn_pattern}, frames within the same generated chunk can attend to different historical frames, indicating that context demand is generated-frame-dependent rather than shared by the whole chunk. Moreover, the same historical frame can receive different attention scores as its relative temporal distance to the current frames changes, suggesting that attention reflects not only content relevance but also temporal position. Therefore, selecting history with a single aggregated attention score may discard frames that are important for specific generated frames, while attention-only ranking can be unstable as relative temporal distance affects attention scores.

Beyond selecting historical frames for each generated frame, fine-grained cache compression also requires allocating appropriate history budgets across attention heads. Existing head-specific strategies mainly rely on aggregated attention scores or heuristic attention patterns, but they do not explicitly estimate head importance for generation quality. As a result, they lack a direct criterion for head-wise KV budget allocation. This motivates assigning larger historical budgets to more important heads.

Taken together, these observations motivate fine-grained KV compression across two dimensions: per-frame history selection and per-head budget allocation. Instead of pruning historical frames with a single aggregated attention score, KV compression should determine \textbf{\emph{which}} historical frames each generated frame should access and \textbf{\emph{how much}} history each attention head should retain. Meanwhile, the selection criterion should go beyond attention alone, since attention scores can be affected by relative temporal distance.

Motivated by this formulation, we propose \textbf{Focused Forcing}, a training-free KV compression method for autoregressive video generation. As shown in Fig.~\ref{fig:teaser}(b), Focused Forcing addresses fine-grained context allocation with three key designs:
\textbf{1) Generated-Frame-Wise History Selection.} Focused Forcing selects historical frames separately for each generated frame, rather than using one shared selection for the whole generated chunk.
\textbf{2) Content-Aware Scoring.} Focused Forcing combines attention relevance with diversity scores of historical frames, preserving distinctive historical frames when attention scores are affected by relative temporal position.
\textbf{3) Head-Wise Budget Allocation.} Focused Forcing estimates head importance by masking each attention head during denoising rollout and measuring the resulting degradation with a distribution-matching loss, then assigns larger KV budgets to more influential heads.

Experiments across multiple autoregressive generation paradigms show that Focused Forcing consistently reduces cache overhead without additional training. As shown in Fig.~\ref{fig:teaser}(c), it achieves up to 1.48$\times$ end-to-end acceleration while also improving visual quality and text alignment, demonstrating that fine-grained KV compression enhances both efficiency and quality in long-horizon video generation.

Our contributions are summarized as follows:
\begin{itemize}[leftmargin=*, labelsep=0.5em, itemsep=5pt, parsep=0pt, topsep=0pt]
    \item We observe that attention-based history selection in autoregressive video generation is generated-frame-dependent and relative-position-sensitive, and that different attention heads have unequal impact on generation quality.
    \item We propose \textbf{Focused Forcing}, a training-free KV compression method that performs fine-grained context allocation through generated-frame-wise history selection, content-aware scoring with attention and diversity scores of historical frames, and head-wise differentiated KV budgets.
    \item Experiments across multiple autoregressive generation paradigms show that Focused Forcing achieves up to \textbf{1.48$\times$} end-to-end acceleration without training, while \textbf{improving visual quality and text alignment}.
\end{itemize}

\section{Related Work}
\label{related_works}

\paragraph{Video Diffusion Models.}
Diffusion models \citep{ddpm,ma2025follow,ma2025controllable,blattmann2023SVD,dit} have set a strong baseline for image and video generation. Benefiting from scaled data and compute, recent video generators achieve high visual fidelity and temporal consistency, e.g., Wan~\citep{wan2_1}, HunyuanVideo 1.5~\citep{hunyuanvideo2025}, LTX Video~\citep{ltx2}, and Seedance2~\citep{seedance2}, most of which rely on bidirectional spatiotemporal attention that ensures global consistency but is costly for long-form or interactive use. To support long-horizon and online generation, causal autoregressive video diffusion synthesizes future frames sequentially with historical context maintained through KV caching, via causal conversion from bidirectional models~\citep{causvid}, rollout-aware training~\citep{self_forcing}, streaming rolling-window generation~\citep{rolling_forcing,zhang2026astrolabe, longlive}, and AR diffusion distillation~\citep{causal_forcing}, and has been scaled in industrial world-modeling and game-content systems~\citep{lingbot-world,Hunyuan-GameCraft-2,helios}.

\paragraph{Efficient Video Generation.}
Efficient video diffusion typically reduces denoising steps, per-step cost, or redundant computation along the trajectory. Better samplers accelerate inference via deterministic non-Markovian sampling~\citep{ddim}, high-order ODE solvers~\citep{lu2022dpmsolverpp}, and unified predictor-corrector schemes~\citep{zhao2023unipc}. Distillation and consistency training compress trajectories into few- or one-step generation~\citep{salimans2022progressive,luo2023latentconsistencymodelssynthesizing,frans2025stepdiffusionshortcutmodels,geng2025meanflowsonestepgenerative}. Per-step cost is cut by memory-efficient or low-bit attention, structured attention for real-time video~\citep{monarch_rt}, and multi-GPU parallelism~\citep{flashattention,flashattention2,SageAttention,SageAttention2,li2024distrifusion,fang2024pipefusion}. Training-free diffusion caching exploits feature redundancy across adjacent denoising steps, evolving from reusing slow features in U-Net/DiT~\citep{deepcache,fora} to finer-grained or video-aware reuse~\citep{toca, teacache}, and to predictive caching that extrapolates future activations~\citep{taylorseer, lesa}.

\paragraph{KV Cache Compression for Autoregressive Generation.}
In autoregressive language models, KV cache compression prunes or allocates history via attention salience, recency, sink tokens, or head-wise specialization~\citep{streamingllm,h2o,duoattention,fastgen}. Video counterparts face growing KV caches and spatiotemporal redundancy. Dummy Forcing~\citep{dummy_forcing} studies head-level dependence on long-range context and applies head-wise KV compression. Relax Forcing~\citep{relax_forcing} organizes temporal memory into Sink, History, and Tail, and dynamically selects informative historical frames. These methods improve cache efficiency and rollout stability, but mainly make head-level or coarse temporal context decisions. Our contribution lies in the joint formulation of query-frame-wise history selection and quality-impact-based head-wise budget allocation, rather than introducing either mechanism independently.

\section{Method}
\label{method}

\begin{figure}[htbp]
    \vspace{-2mm}
    \centering
    \includegraphics[width=\linewidth]{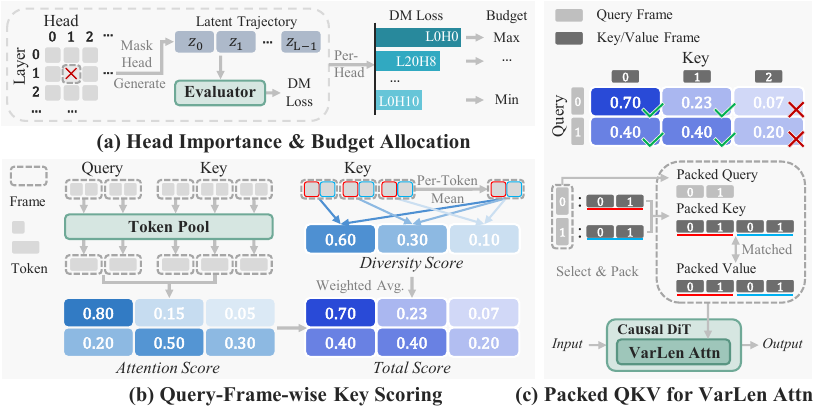}
    \vspace{-6mm}
    \caption{\textbf{Overview of Focused Forcing.} (a) We estimate head importance by masking each head and measuring the DM loss, then allocate larger KV budgets to more important heads. (b) For each query frame and head, we score historical frames by combining attention scores with their diversity scores. (c) The selected QKV rows are packed and computed by variable-length FlashAttention.}
    \label{fig:method}
    \vspace{-2mm}
\end{figure}

\subsection{Preliminary}
\label{sec:preliminary}

\paragraph{Autoregressive Video Diffusion.}
Autoregressive video diffusion generates videos causally, where each frame or chunk is conditioned on previously generated context. Given a video decomposed into autoregressive units
$x^{1:N}=(x^1,x^2,\ldots,x^N)$, the joint distribution is factorized as $p(x^{1:N})=\prod_{i=1}^{N}p(x^i\mid x^{<i})$, where each $x^i$ can denote either a frame or a chunk.

In chunk-wise generation, tokens are grouped into $F_q$ query frames from the current chunk and $F_h$ historical frames from the KV cache:
\begin{equation}
    Q=[Q^{(1)},\ldots,Q^{(F_q)}],
    \qquad
    K=[K^{(1)},\ldots,K^{(F_h)}],
\end{equation}
where $Q^{(f)}$ and $K^{(t)}$ denote tokens from the $f$-th query frame and the $t$-th historical frame.

\paragraph{Distribution Matching Loss.}
Following distribution matching distillation~\cite{dmd,dmd2}, the distribution matching gradient is given by the difference between fake and real score functions. We use its surrogate loss form:
\begin{equation}
    \mathcal{L}_{\mathrm{DM}}(x)
    =
    \frac{1}{2}
    \left\|
    x-\mathrm{sg}\!\left(x-g_{\mathrm{DM}}(x_t,t)\right)
    \right\|_2^2,
    \quad
    g_{\mathrm{DM}}(x_t,t)
    =
    w_t\alpha_t
    \left(
    s_{\mathrm{fake}}(x_t,t)-s_{\mathrm{real}}(x_t,t)
    \right),
\end{equation}
where $x_t$ is obtained by adding noise to $x$, $s_{\mathrm{fake}}$ and $s_{\mathrm{real}}$ denote the score functions of the generated and target distributions, $w_t$ is a timestep-dependent weight, $\alpha_t$ denotes the signal scaling coefficient at timestep $t$, and $\mathrm{sg}(\cdot)$ denotes stop-gradient. A larger $\mathcal{L}_{\mathrm{DM}}$ indicates stronger distributional deviation and is used to measure generation degradation.

\paragraph{Variable-Length FlashAttention.}
For samples with different sequence lengths $\{L_b\}_{b=1}^{B}$, where $Q_b,K_b,V_b \in \mathbb{R}^{1 \times L_b \times H \times D}$, variable-length FlashAttention~\cite{flashattention, flashattention2} concatenates them along the token dimension:
\begin{equation}
    Q_{\mathrm{pack}}=[Q_1;\ldots;Q_B],\quad
    K_{\mathrm{pack}}=[K_1;\ldots;K_B],\quad
    V_{\mathrm{pack}}=[V_1;\ldots;V_B].
\end{equation}
Let $c^q_0=c^k_0=0$, $c^q_b=\sum_{i=1}^{b}L^q_i$, and $c^k_b=\sum_{i=1}^{b}L^k_i$ denote the cumulative boundaries for query and key-value sequences. The attention output is written back to the packed query layout:
\begin{equation}
    O_{\mathrm{pack}}[c^q_{b-1}:c^q_b]
    =
    \mathrm{softmax}
    \left(
        \frac{
            Q_{\mathrm{pack}}[c^q_{b-1}:c^q_b]\,
            K_{\mathrm{pack}}[c^k_{b-1}:c^k_b]^\top
        }{\sqrt{D}}
    \right)
    V_{\mathrm{pack}}[c^k_{b-1}:c^k_b],
\end{equation}
where $O_{\mathrm{pack}}$ denotes the packed attention output, and $D$ is the attention head dimension.

\subsection{Head-Wise Budget Allocation}
\label{sec:head_budget}

Different attention heads can have unequal impact on generation quality, so assigning the same KV budget to all heads may over-compress important heads while wasting cache on less important ones. As shown in Fig.~\ref{fig:method}(a), we estimate head importance by masking each layer-head pair during denoising rollout and measuring the induced DM loss. The head-importance estimation is performed once offline, and the resulting KV budget table is reused during inference.

For a layer-head pair $m=(\ell,h)$, we mask the output of this head while keeping other computations unchanged, obtaining a perturbed latent trajectory $\tilde{\mathbf{z}}^{(m)}$. The importance of head $m$ is measured by the average DM loss over prompts and sampled temporal windows:
\begin{equation}
    I_m
    =
    \frac{1}{|\mathcal{P}|\,|\mathcal{W}|}
    \sum_{p\in\mathcal{P}}
    \sum_{w\in\mathcal{W}}
    \mathcal{L}_{\mathrm{DM}}
    \left(
        \tilde{\mathbf{z}}^{(m)}_{p,w}
    \right),
\end{equation}
where $\mathcal{P}$ denotes the number of prompts, and $\mathcal{W}$ denotes sampled temporal windows from the generated latent trajectory. A larger $I_m$ indicates that masking head $m$ causes stronger generation degradation. The detailed algorithmic procedure is provided in App.~\ref{app:head_importance}.

\begin{figure}[htbp]
    \vspace{-2mm}
    \centering
    \includegraphics[width=0.9\linewidth]{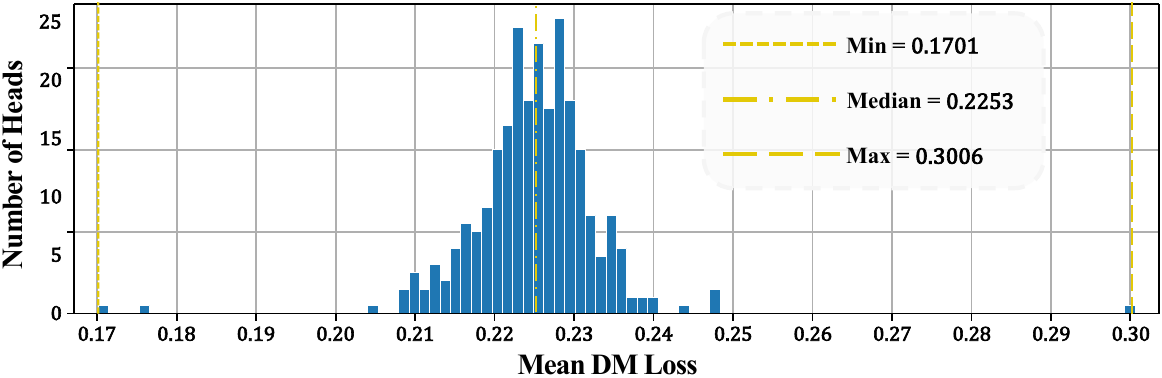}
    \vspace{-2mm}
    \caption{\textbf{Head-wise importance is non-uniform.} Each bar counts the number of heads falling into its DM-loss interval, and the vertical lines mark the minimum, median, and maximum values.}
    \label{fig:dm_loss}
    \vspace{-2mm}
\end{figure}

The DM loss distribution in Fig.~\ref{fig:dm_loss} is clearly non-uniform. For each head, we average the DM loss over sampled temporal windows and prompts, and report the resulting mean DM loss as its importance score. The histogram shows that most heads cluster in a narrow range around the median, while a few heads have much higher or lower mean loss, forming a long-tailed distribution. The mean loss reflects the expected degradation caused by masking a head and is used as the head-importance score for budget allocation.

After obtaining $\{I_m\}$ for all layer-head pairs, we normalize the importance scores:
\begin{equation}
    \hat{I}_m
    =
    \frac{
        I_m-I_{\min}
    }{
        I_{\max}-I_{\min}+\epsilon
    }.
\end{equation}
We then map each normalized score to a discrete KV budget:
\begin{equation}
    b_m
    =
    \mathrm{round}
    \left(
        b_{\min}
        +
        \hat{I}_m^{\gamma}
        \left(
            b_{\max}-b_{\min}
        \right)
    \right),
\end{equation}
where $b_{\min}$ and $b_{\max}$ are the minimum and maximum budgets, and $\gamma$ controls the curvature of the mapping. More important heads therefore retain more historical frames.

\subsection{Query-Frame-Wise History Scoring and Selection}
\label{sec:frame_selection}

Frames within the same generated chunk may require different historical context. Focused Forcing therefore selects historical frames separately for each query frame and head by combining attention scores with diversity scores under the assigned head-wise budget, as shown in Fig.~\ref{fig:method}(b).
\paragraph{Attention Scores.}
For efficiency, we group tokens within each frame into $P$ groups. The grouped query and key features are represented as
$\bar{Q}\in\mathbb{R}^{B\times QF\times P\times H\times D}$ and
$\bar{K}\in\mathbb{R}^{B\times KF\times P\times H\times D}$.
Let $\bar{Q}_{b,q_f,u,h}$ and $\bar{K}_{b,k_f,v,h}$ denote the pooled features for query frame $q_f$, historical frame $k_f$, group indices $u,v$, and head $h$. The frame-level attention score is defined as:
\begin{equation}
    A_{b,q_f,h,k_f}
    =
    \frac{1}{P^2}
    \sum_{u=1}^{P}
    \sum_{v=1}^{P}
    \frac{
        \left\langle
        \bar{Q}_{b,q_f,u,h},
        \bar{K}_{b,k_f,v,h}
        \right\rangle
    }{\sqrt{D}},
\end{equation}
where $D$ is the head dimension. We standardize $A_{b,q_f,h,k_f}$ along the historical-frame dimension for each batch, query frame, and head, so that scores over candidate historical frames have zero mean and unit variance, yielding $\tilde{A}_{b,q_f,h,k_f}$.

\paragraph{Diversity Scores of Historical Frames.}
Attention scores may not fully capture frame distinctiveness, especially when they vary with relative temporal distance, as proved in App.~\ref{app:temporal_rope}. We therefore compute diversity scores from historical key frames. Let $K^{(k_f)} \in \mathbb{R}^{B \times L \times H \times D}$ denote the key of the $k_f$-th historical frame, where $L$ is the number of tokens per frame. For each batch index $b$, token index $l$, and head index $h$, we compute the average key representation over historical frames and measure the redundancy of historical frame $k_f$ by its average cosine similarity to this representation:
\begin{equation}
    \bar{K}^{\mathrm{mean}}_{b,l,h}
    =
    \frac{1}{F_h}
    \sum_{k_f=1}^{F_h}
    K^{(k_f)}_{b,l,h},
    \quad
    R_{b,h,k_f}
    =
    \frac{1}{L}
    \sum_{l=1}^{L}
    \left\langle
        \frac{K^{(k_f)}_{b,l,h}}{\left\|K^{(k_f)}_{b,l,h}\right\|_2+\epsilon},
        \frac{\bar{K}^{\mathrm{mean}}_{b,l,h}}{\left\|\bar{K}^{\mathrm{mean}}_{b,l,h}\right\|_2+\epsilon}
    \right\rangle .
\end{equation}
Here, $\epsilon$ is a small constant for numerical stability. A historical frame is less distinctive if it is closer to the average historical representation. We therefore define the diversity score of historical frame $k_f$ as negative redundancy:
\begin{equation}
    D_{b,h,k_f}=-R_{b,h,k_f}.
\end{equation}
After standardization along the historical-frame dimension, the diversity score of each historical frame is broadcast to each query frame, yielding $\tilde{D}_{b,q_f,h,k_f}$.

\begin{figure}[htbp]
    \vspace{-2mm}
    \centering
    \includegraphics[width=\linewidth]{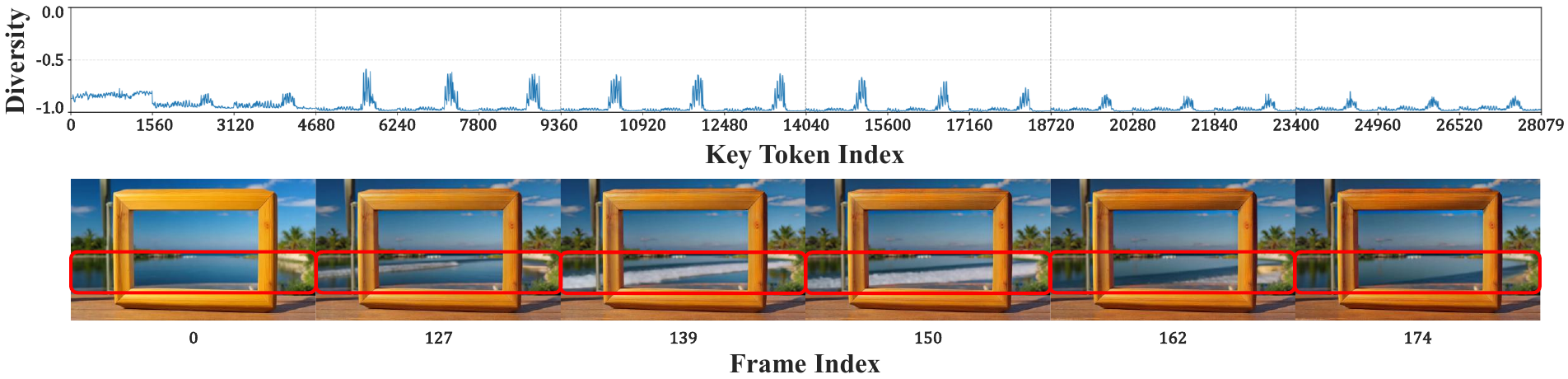}
    \vspace{-6mm}
    \caption{
    \textbf{Diversity scores highlight changing regions.}
    Historical frames with high diversity scores often contain regions that differ from the average historical representation.
    }
    \label{fig:key_diversity_pattern_0}
    \vspace{-2mm}
\end{figure}

As shown in Fig.~\ref{fig:key_diversity_pattern_0}, high diversity scores highlight regions that differ across historical frames, while static regions stay close to the average representation. Thus, diversity scores complement attention scores by preserving distinctive historical frames that may be under-ranked by attention alone.

\paragraph{Final Score and Selection.}
We compute the final score by weighting the attention score and the diversity score of each historical frame, and select historical frames under the assigned budget:
\begin{equation}
    S_{b,q_f,h,k_f}
    =
    \lambda \tilde{A}_{b,q_f,h,k_f}
    +
    (1-\lambda)\tilde{D}_{b,q_f,h,k_f},
    \quad
    \mathcal{S}_{b,q_f,h}
    =
    \mathrm{TopK}_{k_f}
    \left(
        S_{b,q_f,h,k_f},
        b_{\ell,h}
    \right).
\end{equation}
where $\lambda$ controls the trade-off between the two scores, and $b_{\ell,h}$ is the budget of the current layer-head pair. In practice, we reserve anchor and generated frames, and use the remaining budget for the highest-scoring historical frames to preserve reference and consistency.

\subsection{Packed QKV for Variable-Length FlashAttention}
\label{sec:varlen_attn}

Because the selected historical frames vary across query frames and heads, the resulting KV lengths are irregular. As shown in Fig.~\ref{fig:method}(c), we pack the corresponding QKV rows into variable-length sequences. For each query frame and head, the query length is fixed, while the key-value length is determined by the assigned budget. We compute attention only over the retained historical frames:
\begin{equation}
    O_{\mathrm{pack}}
    =
    \mathrm{VarLenFlashAttn}
    \left(
        Q_{\mathrm{pack}},
        K_{\mathrm{pack}},
        V_{\mathrm{pack}},
        \mathrm{cu\_q},
        \mathrm{cu\_k}
    \right),
\end{equation}
Specifically, each batch, query-frame, and head tuple forms an independent variable-length attention sequence. The selected K/V frame blocks are gathered using identical indices and concatenated into packed tensors, with \texttt{cu\_q} and \texttt{cu\_k} recording the sequence boundaries. Variable-length FlashAttention processes these sequences independently, and the outputs are scattered back to their original query positions. This execution preserves the selection across query frames and heads while avoiding dense attention.

\section{Experiments}
\label{experiments}

\subsection{Settings}

\paragraph{Baselines and comparisons.}
\label{exp:baselines}
We evaluate Focused Forcing under two settings. To verify compatibility across autoregressive paradigms, we plug it into Self Forcing~\cite{self_forcing}, LongLive~\cite{longlive}, Rolling Forcing~\cite{rolling_forcing}, and Causal Forcing~\cite{causal_forcing}, comparing against their original baselines as well as representative autoregressive diffusion models NOVA~\cite{nova}, SkyReels-V2~\cite{skyreels_v2}, and MAGI-1~\cite{magi_1}. For acceleration, we take chunk-wise Self Forcing as the base model and compare with its Attention Sink variant, MonarchRT~\cite{monarch_rt}, TaylorSeer~\cite{taylorseer}, and Dummy Forcing~\cite{dummy_forcing}, which respectively represent structured sparse attention, cached-feature-based prediction, and head-wise KV cache pruning. Implementation details are provided in App.~\ref{app:implementation}.

\paragraph{Evaluation.}
\label{exp:evaluations}
Following prior work~\cite{self_forcing++}, we evaluate long video generation on VBench-Long~\cite{vbench++} using the first 128 prompts from MovieGen~\cite{moviegen} refined by Qwen2.5-7B-Instruct~\cite{qwen3}. We follow the VBench-Long protocol to assess generation quality, and further evaluate inference efficiency using the average generation latency, average self-attention latency, and their corresponding speedups. Detailed metric definitions are provided in App.~\ref{app:metrics}.

\subsection{Comparison with Autoregressive Video Diffusion Models}

\paragraph{Quantitative results.}
We compare Focused Forcing with state-of-the-art autoregressive video models on VBench-Long. As shown in Tab.~\ref{tab:quality_comparison}, Focused Forcing consistently improves the efficiency-quality trade-off across different autoregressive paradigms. On Self Forcing, it achieves 1.45$\times$ end-to-end speedup, improves visual quality from 76.58 to 80.00, and substantially increases Dynamic Degree from 41.55 to 61.39. Similar improvements on LongLive, Rolling Forcing, and Causal Forcing show that Focused Forcing generalizes across different autoregressive backbones.

\begin{table}[H]
    \centering
    \caption{Quantitative comparison for long video generation on VBench-Long.}
    \vspace{-1.5mm}
    \resizebox{\columnwidth}{!}{
        \setlength{\tabcolsep}{3pt}
        \renewcommand{\arraystretch}{1.1}
        \begin{tabular}{@{}l | c c | c c c c c c | c c@{}}
            \toprule
            \multirow{2}{*}{\textbf{Methods}} &
            \multirow{2}{*}{\makecell{\textbf{Gen.}\\\textbf{Latency/s}}} &
            \multirow{2}{*}{\makecell{\textbf{Gen.}\\\textbf{Speedup}}} &
            \multirow{2}{*}{\makecell{\textbf{Subject}\\\textbf{Consistency}}} &
            \multirow{2}{*}{\makecell{\textbf{Background}\\\textbf{Consistency}}} &
            \multirow{2}{*}{\makecell{\textbf{Motion}\\\textbf{Smoothness}}} &
            \multirow{2}{*}{\makecell{\textbf{Dynamic}\\\textbf{Degree}}} &
            \multirow{2}{*}{\makecell{\textbf{Aesthetic}\\\textbf{Quality}}} &
            \multirow{2}{*}{\makecell{\textbf{Imaging}\\\textbf{Quality}}} &
            \multirow{2}{*}{\makecell{\textbf{Visual}\\\textbf{Quality}}} &
            \multirow{2}{*}{\makecell{\textbf{Text}\\\textbf{Alignment}}} \\
            & & & & & & & & & & \\
            \midrule
            \rowcolor{gray!30}
            \multicolumn{11}{l}{\textit{Autoregressive Diffusion Models}}\\
            NOVA
            & 2558.21 & 1.00$\times$ & 90.15 & 93.35 & 98.81 & 34.08 & 44.14 & 38.81 & 66.56 & 24.63\\
            SkyReels-V2
            & 1244.40 & 1.00$\times$ & 96.93 & 96.15 & 98.68 & 38.22 & 58.86 & 63.42 & 75.38 & 27.54\\
            MAGI-1
            & 6740.65 & 1.00$\times$ & 97.29 & 96.53 & 99.10 & 39.10 & 57.93 & 59.24 & 74.87 & 29.41\\
            \midrule
            \rowcolor{gray!30}
            \multicolumn{11}{l}{\textit{Distilled Autoregressive Diffusion Models}}\\
            CausVid
            & 61.69 & 1.00$\times$ & 97.88 & 96.82 & 98.08 & 48.61 & 59.94 & 65.79 & 77.85 & 27.19\\
            Self Forcing
            & 78.06 & 1.00$\times$ & 96.61 & 96.17 & 98.27 & 41.55 & 58.83 & 68.04 & 76.58 & 28.03\\
            \rowcolor{gray!15}
            \quad \textbf{+ Ours}
            & 53.90 & 1.45$\times$ & 95.84 & 95.49 & 98.04 & 61.39 & 60.59 & 68.62 & 80.00 & 28.75\\
            LongLive
            & 90.06 & 1.00$\times$ & 97.60 & 96.36 & 98.90 & 35.42 & 60.74 & 68.06 & 76.18 & 29.06\\
            \rowcolor{gray!15}
            \quad \textbf{+ Ours}
            & 68.60 & 1.31$\times$ & 96.26 & 95.79 & 98.30 & 46.03 & 61.51 & 69.58 & 77.91 & 29.40\\
            Rolling Forcing
            & 83.35 & 1.00$\times$ & 97.88 & 96.47 & 98.77 & 29.67 & 60.69 & 70.76 & 75.71 & 28.85\\
            \rowcolor{gray!15}
            \quad \textbf{+ Ours}
            & 58.20 & 1.43$\times$ & 96.21 & 95.40 & 97.91 & 50.03 & 59.77 & 68.47 & 77.97 & 28.99\\
            Causal Forcing
            & 77.99 & 1.00$\times$ & 95.66 & 94.34 & 97.56 & 94.05 & 56.46 & 68.35 & 84.40 & 26.80\\
            \rowcolor{gray!15}
            \quad \textbf{+ Ours}
            & 52.68 & 1.48$\times$ & 94.53 & 94.11 & 96.63 & 97.77 & 57.54 & 68.29 & 84.81 & 28.27\\
            \bottomrule
        \end{tabular}
    }
    \label{tab:quality_comparison}
    \vspace{-1mm}
\end{table}

\begin{figure}[htbp]
    \vspace{-2mm}
    \centering
    \includegraphics[width=\linewidth]{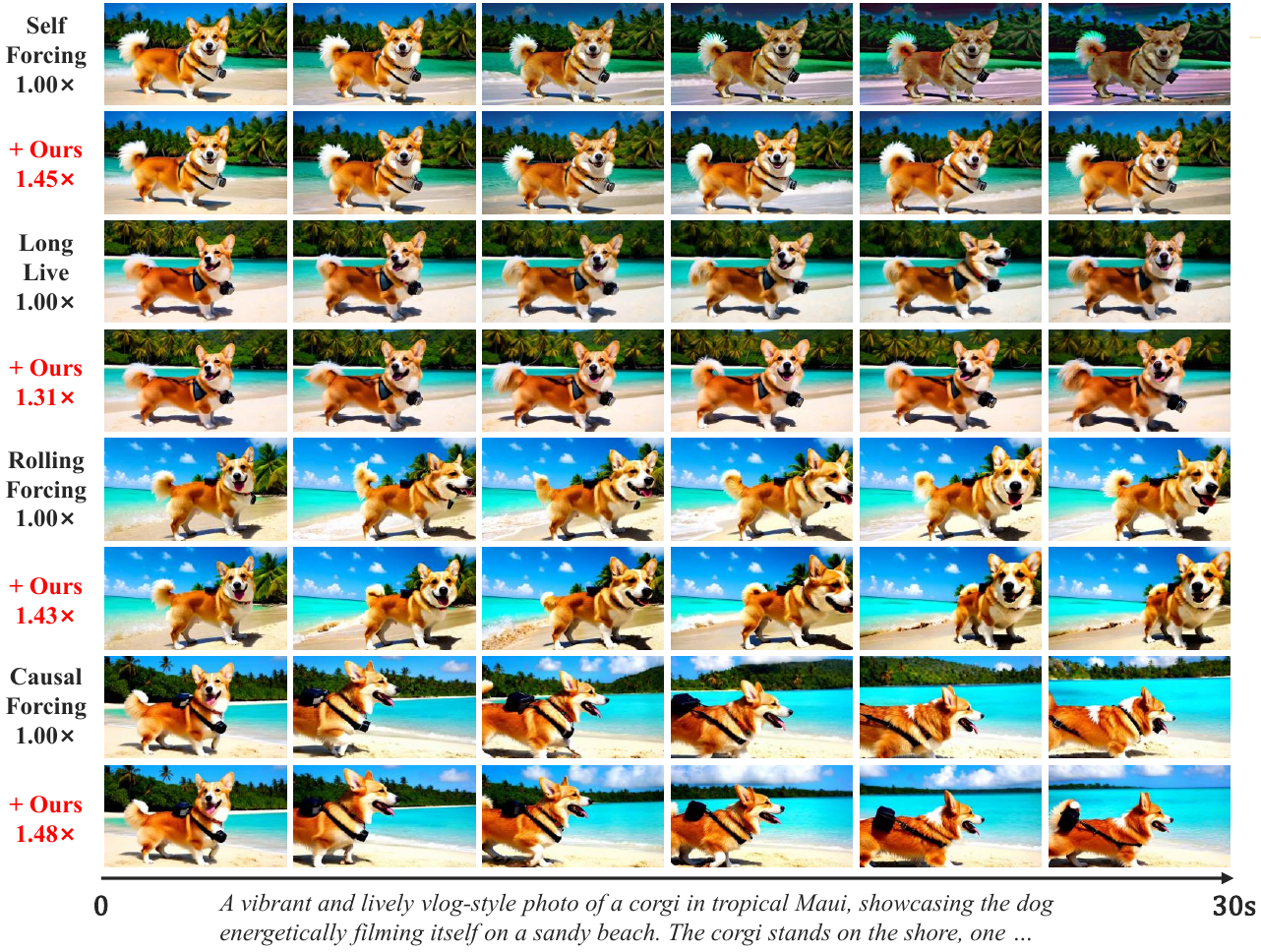}
    \vspace{-6mm}
    \caption{Qualitative comparison on 30s video generation across different autoregressive paradigms.}
    \label{fig:quality_comparison_4}
    \vspace{-4mm}
\end{figure}

\paragraph{Qualitative results.}
As shown in Fig.~\ref{fig:quality_comparison_4}, Self Forcing suffers from noticeable long-horizon degradation, with the dog appearance and background details becoming increasingly unstable over time. Although Long-Live better preserves the overall scene layout, it exhibits evident color degradation in later frames. Focused Forcing leads to more stable visual results in these cases, with the corgi's appearance and background better preserved. For Rolling Forcing and Causal Forcing, whose baselines already produce relatively stable results, the visual improvement is less pronounced. Nevertheless, Focused Forcing preserves comparable visual quality while substantially improving inference efficiency, indicating its effectiveness as a general acceleration module for long-video generation.

\subsection{Comparison with Acceleration Methods}

\paragraph{Quantitative results.}
We further compare Focused Forcing with representative acceleration methods on chunk-wise Self Forcing. Shown in Tab.~\ref{tab:efficiency_comparison}, Focused Forcing achieves 2.77$\times$ attention speedup and 1.45$\times$ end-to-end speedup, while delivering the best visual quality among all acceleration methods. It improves visual quality over Attention Sink from 79.2 to 80.0 with substantial speedup, surpasses MonarchRT and TaylorSeer in both quality and end-to-end speedup, and remains competitive with Dummy Forcing in efficiency while achieving higher visual quality. Fig.~\ref{fig:radar} further highlights its strong Dynamic Degree and competitive performance across other quality dimensions, confirming a stronger speed-quality trade-off through selective KV-context compression.

\begin{table}[H]
    \centering
    \caption{Efficiency comparison for long video generation on VBench-Long.}
    \vspace{-1mm}
    \resizebox{0.85\columnwidth}{!}{
        \setlength{\tabcolsep}{3pt}
        \renewcommand{\arraystretch}{1.1}
        \begin{tabular}{@{}l | c c c c | c c | c c@{}}
            \toprule
            \multirow{2}{*}{\textbf{Methods}} &
            \multirow{2}{*}{\makecell{\textbf{Attn.}\\\textbf{Latency/s}}} &
            \multirow{2}{*}{\makecell{\textbf{Attn.}\\\textbf{Speedup}}} &
            \multirow{2}{*}{\makecell{\textbf{Gen.}\\\textbf{Latency/s}}} &
            \multirow{2}{*}{\makecell{\textbf{Gen.}\\\textbf{Speedup}}} &
            \multirow{2}{*}{\makecell{\textbf{Temporal}\\\textbf{Quality}}} &
            \multirow{2}{*}{\makecell{\textbf{Frame-wise}\\\textbf{Quality}}} &
            \multirow{2}{*}{\makecell{\textbf{Visual}\\\textbf{Quality}}} &
            \multirow{2}{*}{\makecell{\textbf{Text}\\\textbf{Alignment}}} \\
            & & & & & & & & \\
            \midrule
            Self Forcing
            & 37.29 & 1.00$\times$ & 78.06 & 1.00$\times$ & 83.15 & 63.44 & 76.58 & 28.03 \\
            + Attention Sink
            & 37.15 & 1.00$\times$ & 78.07 & 1.00$\times$ & 86.46 & 64.68 & 79.20 & 28.42 \\
            MonarchRT
            & 33.13 & 1.13$\times$ & 72.61 & 1.08$\times$ & 85.95 & 64.04 & 78.65 & 29.24 \\
            TaylorSeer
            & 29.77 & 1.25$\times$ & 68.88 & 1.13$\times$ & 85.67 & 64.38 & 78.57 & 28.85 \\
            Dummy Forcing
            & 13.35 & 2.79$\times$ & 53.64 & 1.46$\times$ & 84.79 & 65.55 & 78.38 & 28.57 \\
            \rowcolor{gray!15}
            \textbf{Ours}
            & 13.48 & 2.77$\times$ & 53.90 & 1.45$\times$ & 87.69 & 64.61 & 80.00 & 28.75 \\
            \bottomrule
        \end{tabular}
    }
    \label{tab:efficiency_comparison}
    \vspace{-1mm}
\end{table}

\paragraph{Qualitative results.}

\begin{figure}[htbp]
    \vspace{-2mm}
    \centering
    \includegraphics[width=\linewidth]{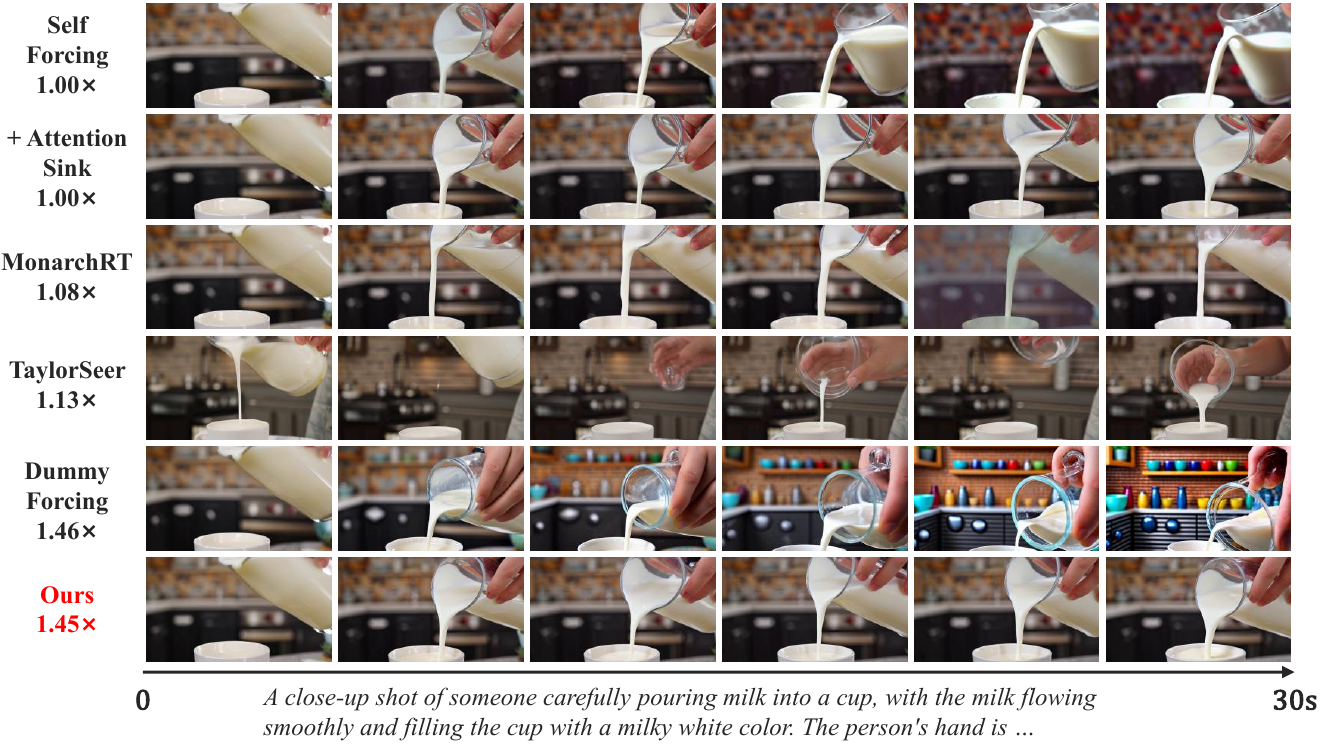}
    \vspace{-6mm}
    \caption{Qualitative comparison of inference acceleration methods on 30s video generation.}
    \label{fig:efficiency_comparison_0}
    \vspace{-4mm}
\end{figure}

As shown in Fig.~\ref{fig:efficiency_comparison_0}, Attention Sink preserves the overall scene layout, whereas Self Forcing and TaylorSeer both exhibit glass-shape drift and physically implausible milk-flow positions, and MonarchRT and Dummy Forcing introduce visible artifacts and quality degradation, respectively. In contrast, Focused Forcing yields a more coherent trajectory, with the object stream evolving smoothly throughout the rollout, indicating that our method removes redundant historical information while preserving motion-relevant temporal cues.

\subsection{Ablation Studies}
We ablate the KV budget and attention weight, with detailed results provided in App.~\ref{app:ablation}.

\begin{figure}[htbp]
    \centering
    \begin{minipage}{0.315\linewidth}
        \centering
        \includegraphics[width=\linewidth]{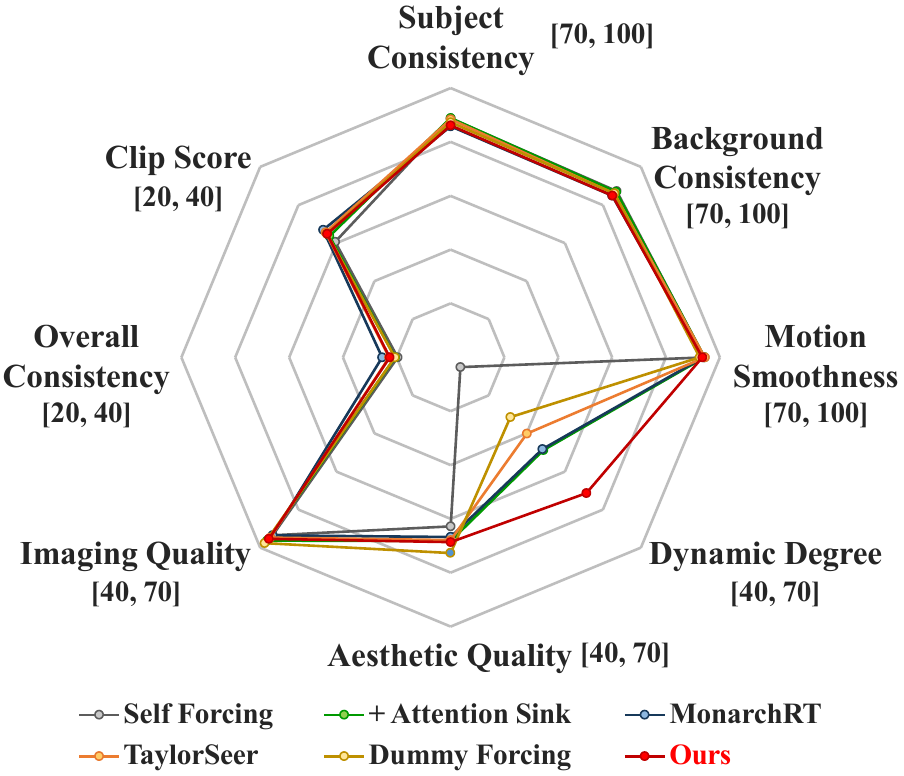}
        \caption{
        Metric comparison\\with acceleration methods.
        }
        \label{fig:radar}
    \end{minipage}
    \hfill
    \begin{minipage}{0.315\linewidth}
        \centering
        \includegraphics[width=\linewidth]{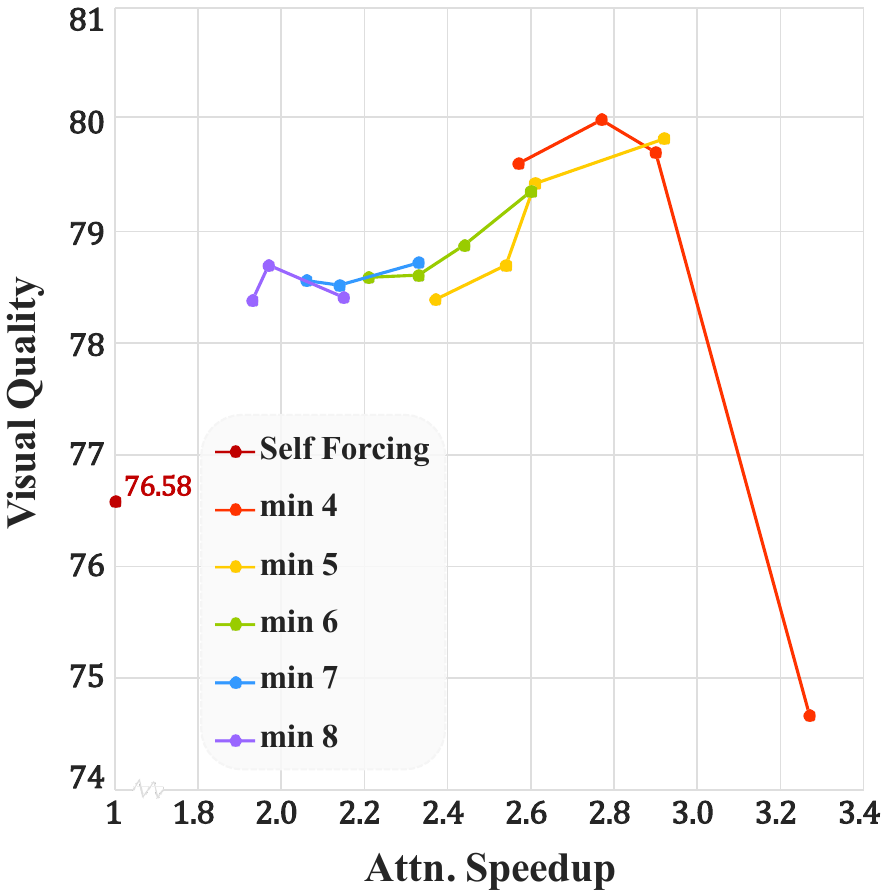}
        \vspace{-6mm}
        \caption{
        Ablation study on the KV budget.
        }
        \label{fig:ablation_kv_budget}
    \end{minipage}
    \hfill
    \begin{minipage}{0.315\linewidth}
        \centering
        \includegraphics[width=\linewidth]{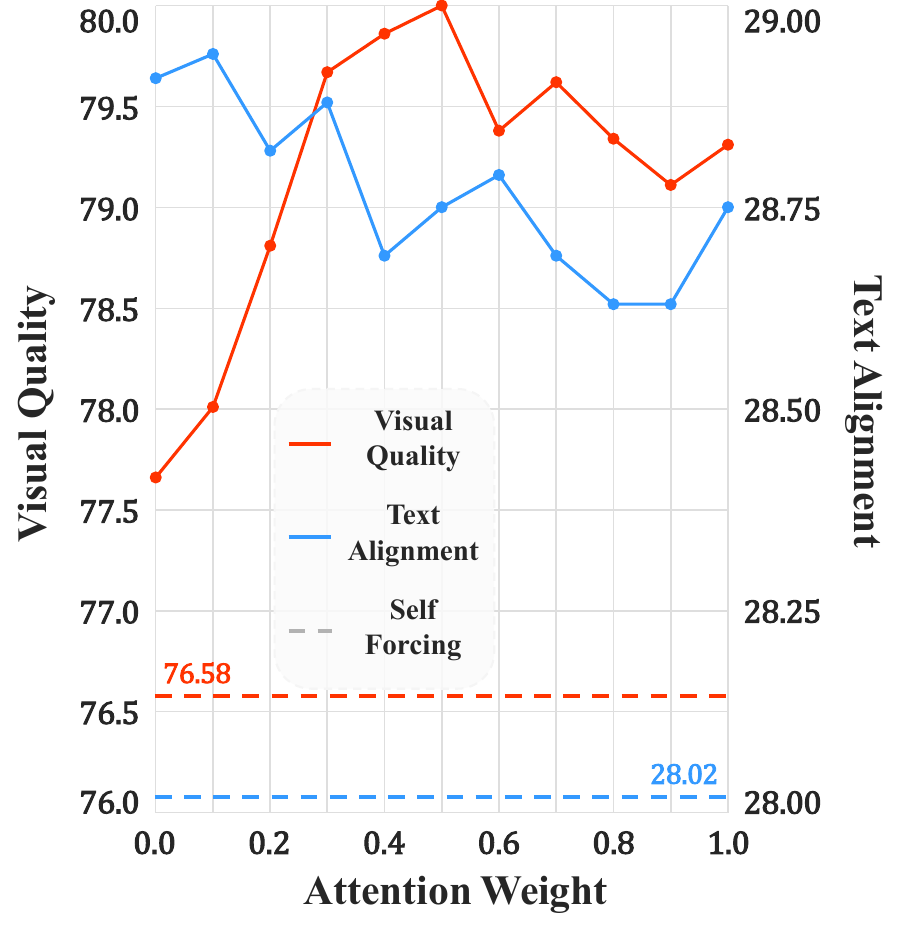}
        \vspace{-6mm}
        \caption{
        Ablation study on the attention weight.
        }
        \label{fig:ablation_attn_weight}
    \end{minipage}
    \vspace{-2mm}
\end{figure}

\paragraph{Ablation on KV budget.}
As shown in Fig.~\ref{fig:ablation_kv_budget}, Self Forcing is the uncompressed baseline. 
For Focused Forcing, each colored curve represents a fixed minimum KV budget $b_{\min}$, as indicated by the legend, e.g., ``min 4'' denotes $b_{\min}=4$. The markers on the same curve correspond to different maximum KV budgets $b_{\max}$, which are decreased from 15 with a step size of 3. Moving along a curve toward larger attention speedup therefore indicates stronger KV compression. The results show that moderate budget settings achieve the best trade-off, substantially improving attention speedup over Self Forcing while maintaining high visual quality, whereas overly aggressive compression leads to noticeable quality degradation.

\paragraph{Ablation on attention weight.}
As shown in Fig.~\ref{fig:ablation_attn_weight}, all settings outperform Self Forcing in both visual quality and text alignment. Visual quality improves as the attention weight increases and peaks at a moderate value, while larger weights slightly reduce both metrics. This indicates that combining attention with diversity is important for selecting useful yet non-redundant historical context.

\section{Conclusion}
\label{conclusion}
In this paper, we present \textbf{Focused Forcing}, a training-free KV compression method for autoregressive video generation. By reformulating KV compression as fine-grained context allocation, Focused Forcing selects historical frames separately for each query frame, combines attention scores with diversity scores of historical frames, and assigns differentiated KV budgets according to head importance. This design preserves relevant and distinctive historical context under compact KV budgets. Experiments on VBench-Long across multiple autoregressive paradigms show that Focused Forcing consistently improves the efficiency-quality trade-off, achieving up to \textbf{1.48$\times$} end-to-end acceleration without training while \textbf{improving visual quality and text alignment}. These results show that fine-grained and content-aware KV selection is an effective strategy for scalable long-horizon autoregressive video generation.


\bibliographystyle{plainnat}
\bibliography{neurips_2026}

@inproceedings{dmd,
    author    = {Yin, Tianwei and Gharbi, Micha\"el and Zhang, Richard and Shechtman, Eli and Durand, Fr\'edo and Freeman, William T. and Park, Taesung},
    title     = {One-step Diffusion with Distribution Matching Distillation},
    booktitle = {Proceedings of the IEEE/CVF Conference on Computer Vision and Pattern Recognition (CVPR)},
    month     = {June},
    year      = {2024},
    pages     = {6613-6623}
}

@inproceedings{dmd2,
    author = {Yin, Tianwei and Gharbi, Micha\"{e}l and Park, Taesung and Zhang, Richard and Shechtman, Eli and Durand, Fr\'{e}do and Freeman, William T.},
    booktitle = {Advances in Neural Information Processing Systems},
    doi = {10.52202/079017-1505},
    editor = {A. Globerson and L. Mackey and D. Belgrave and A. Fan and U. Paquet and J. Tomczak and C. Zhang},
    pages = {47455--47487},
    publisher = {Curran Associates, Inc.},
    title = {Improved Distribution Matching Distillation for Fast Image Synthesis},
    url = {https://proceedings.neurips.cc/paper_files/paper/2024/file/54dcf25318f9de5a7a01f0a4125c541e-Paper-Conference.pdf},
    volume = {37},
    year = {2024}
}

@inproceedings{cfg,
    title={Classifier-Free Diffusion Guidance},
    author={Jonathan Ho and Tim Salimans},
    booktitle={NeurIPS 2021 Workshop on Deep Generative Models and Downstream Applications},
    year={2021},
    url={https://openreview.net/forum?id=qw8AKxfYbI}
}

@inproceedings{flashattention,
    author = {Dao, Tri and Fu, Dan and Ermon, Stefano and Rudra, Atri and R\'{e}, Christopher},
    booktitle = {Advances in Neural Information Processing Systems},
    editor = {S. Koyejo and S. Mohamed and A. Agarwal and D. Belgrave and K. Cho and A. Oh},
    pages = {16344--16359},
    publisher = {Curran Associates, Inc.},
    title = {FlashAttention: Fast and Memory-Efficient Exact Attention with IO-Awareness},
    url = {https://proceedings.neurips.cc/paper_files/paper/2022/file/67d57c32e20fd0a7a302cb81d36e40d5-Paper-Conference.pdf},
    volume = {35},
    year = {2022}
}

@inproceedings{flashattention2,
    author = {Dao, Tri},
    booktitle = {International Conference on Learning Representations},
    editor = {B. Kim and Y. Yue and S. Chaudhuri and K. Fragkiadaki and M. Khan and Y. Sun},
    pages = {35549--35562},
    title = {FlashAttention-2: Faster Attention with Better Parallelism and Work Partitioning},
    url = {https://proceedings.iclr.cc/paper_files/paper/2024/file/98ed250b203d1ac6b24bbcf263e3d4a7-Paper-Conference.pdf},
    volume = {2024},
    year = {2024}
}

@inproceedings{nova,
    title={Autoregressive Video Generation without Vector Quantization},
    author={Haoge Deng and Ting Pan and Haiwen Diao and Zhengxiong Luo and Yufeng Cui and Huchuan Lu and Shiguang Shan and Yonggang Qi and Xinlong Wang},
    booktitle={The Thirteenth International Conference on Learning Representations},
    year={2025},
    url={https://openreview.net/forum?id=JE9tCwe3lp}
}

@misc{magi_1,
    title={MAGI-1: Autoregressive Video Generation at Scale}, 
    author={Sand. ai and Hansi Teng and Hongyu Jia and Lei Sun and Lingzhi Li and Maolin Li and Mingqiu Tang and Shuai Han and Tianning Zhang and W. Q. Zhang and Weifeng Luo and Xiaoyang Kang and Yuchen Sun and Yue Cao and Yunpeng Huang and Yutong Lin and Yuxin Fang and Zewei Tao and Zheng Zhang and Zhongshu Wang and Zixun Liu and Dai Shi and Guoli Su and Hanwen Sun and Hong Pan and Jie Wang and Jiexin Sheng and Min Cui and Min Hu and Ming Yan and Shucheng Yin and Siran Zhang and Tingting Liu and Xianping Yin and Xiaoyu Yang and Xin Song and Xuan Hu and Yankai Zhang and Yuqiao Li},
    year={2025},
    eprint={2505.13211},
    archivePrefix={arXiv},
    primaryClass={cs.CV},
    url={https://arxiv.org/abs/2505.13211}, 
}

@misc{skyreels_v2,
    title={SkyReels-V2: Infinite-length Film Generative Model}, 
    author={Guibin Chen and Dixuan Lin and Jiangping Yang and Chunze Lin and Junchen Zhu and Mingyuan Fan and Hao Zhang and Sheng Chen and Zheng Chen and Chengcheng Ma and Weiming Xiong and Wei Wang and Nuo Pang and Kang Kang and Zhiheng Xu and Yuzhe Jin and Yupeng Liang and Yubing Song and Peng Zhao and Boyuan Xu and Di Qiu and Debang Li and Zhengcong Fei and Yang Li and Yahui Zhou},
    year={2025},
    eprint={2504.13074},
    archivePrefix={arXiv},
    primaryClass={cs.CV},
    url={https://arxiv.org/abs/2504.13074}, 
}

@article{wan2_1,
    title={Wan: Open and Advanced Large-Scale Video Generative Models}, 
    author={Team Wan and Ang Wang and Baole Ai and Bin Wen and Chaojie Mao and Chen-Wei Xie and Di Chen and Feiwu Yu and Haiming Zhao and Jianxiao Yang and Jianyuan Zeng and Jiayu Wang and Jingfeng Zhang and Jingren Zhou and Jinkai Wang and Jixuan Chen and Kai Zhu and Kang Zhao and Keyu Yan and Lianghua Huang and Mengyang Feng and Ningyi Zhang and Pandeng Li and Pingyu Wu and Ruihang Chu and Ruili Feng and Shiwei Zhang and Siyang Sun and Tao Fang and Tianxing Wang and Tianyi Gui and Tingyu Weng and Tong Shen and Wei Lin and Wei Wang and Wei Wang and Wenmeng Zhou and Wente Wang and Wenting Shen and Wenyuan Yu and Xianzhong Shi and Xiaoming Huang and Xin Xu and Yan Kou and Yangyu Lv and Yifei Li and Yijing Liu and Yiming Wang and Yingya Zhang and Yitong Huang and Yong Li and You Wu and Yu Liu and Yulin Pan and Yun Zheng and Yuntao Hong and Yupeng Shi and Yutong Feng and Zeyinzi Jiang and Zhen Han and Zhi-Fan Wu and Ziyu Liu},
    journal = {arXiv preprint arXiv:2503.20314},
    year={2025}
}

@InProceedings{causvid,
    author    = {Yin, Tianwei and Zhang, Qiang and Zhang, Richard and Freeman, William T. and Durand, Fredo and Shechtman, Eli and Huang, Xun},
    title     = {From Slow Bidirectional to Fast Autoregressive Video Diffusion Models},
    booktitle = {Proceedings of the IEEE/CVF Conference on Computer Vision and Pattern Recognition (CVPR)},
    month     = {June},
    year      = {2025},
    pages     = {22963-22974}
}

@inproceedings{self_forcing,
    title={Self Forcing: Bridging the Train-Test Gap in Autoregressive Video Diffusion},
    author={Xun Huang and Zhengqi Li and Guande He and Mingyuan Zhou and Eli Shechtman},
    booktitle={The Thirty-ninth Annual Conference on Neural Information Processing Systems},
    year={2026},
    url={https://openreview.net/forum?id=mSiN7i0BYH}
}

@inproceedings{rolling_forcing,
    title={Rolling Forcing: Autoregressive Long Video Diffusion in Real Time},
    author={Kunhao Liu and Wenbo Hu and Jiale Xu and Ying Shan and Shijian Lu},
    booktitle={The Fourteenth International Conference on Learning Representations},
    year={2026},
    url={https://openreview.net/forum?id=IAyzXjbfwo}
}

@inproceedings{longlive,
    title={LongLive: Real-time Interactive Long Video Generation},
    author={Shuai Yang and Wei Huang and Ruihang Chu and Yicheng Xiao and Yuyang Zhao and Xianbang Wang and Muyang Li and Enze Xie and Ying-Cong Chen and Yao Lu and Song Han and Yukang Chen},
    booktitle={The Fourteenth International Conference on Learning Representations},
    year={2026},
    url={https://openreview.net/forum?id=nCAODkpsPJ}
}

@article{causal_forcing,
    title={Causal Forcing: Autoregressive Diffusion Distillation Done Right for High-Quality Real-Time Interactive Video Generation},
    author={Zhu, Hongzhou and Zhao, Min and He, Guande and Su, Hang and Li, Chongxuan and Zhu, Jun},
    journal={arXiv preprint arXiv:2602.02214},
    year={2026}
}

@inproceedings{ma2024follow,
  title={Follow your pose: Pose-guided text-to-video generation using pose-free videos},
  author={Ma, Yue and He, Yingqing and Cun, Xiaodong and Wang, Xintao and Chen, Siran and Li, Xiu and Chen, Qifeng},
  booktitle={Proceedings of the AAAI Conference on Artificial Intelligence},
  volume={38},
  number={5},
  pages={4117--4125},
  year={2024}
}

@article{deep_forcing,
    title={Deep Forcing: Training-Free Long Video Generation with Deep Sink and Participative Compression},
    author={Yi, Jung and Jang, Wooseok and Cho, Paul Hyunbin and Nam, Jisu and Yoon, Heeji and Kim, Seungryong},
    journal={arXiv preprint arXiv:2512.05081},
    year={2025}
}

@article{dummy_forcing,
    title={Efficient Autoregressive Video Diffusion with Dummy Head},
    author={Guo, Hang and Jia, Zhaoyang and Li, Jiahao and Li, Bin and Cai, Yuanhao and Wang, Jiangshan and Li, Yawei and Lu, Yan},
    journal={arXiv preprint arXiv:2601.20499},
    year={2026}
}

@article{zhang2026astrolabe,
  title={Astrolabe: Steering Forward-Process Reinforcement Learning for Distilled Autoregressive Video Models},
  author={Zhang, Songchun and Xue, Zeyue and Fu, Siming and Huang, Jie and Kong, Xianghao and Ma, Y and Huang, Haoyang and Duan, Nan and Rao, Anyi},
  journal={arXiv preprint arXiv:2603.17051},
  year={2026}
}

@article{ma2025controllable,
  title={Controllable video generation: A survey},
  author={Ma, Yue and Feng, Kunyu and Hu, Zhongyuan and Wang, Xinyu and Wang, Yucheng and Zheng, Mingzhe and Wang, Bingyuan and Wang, Qinghe and He, Xuanhua and Wang, Hongfa and others},
  journal={arXiv preprint arXiv:2507.16869},
  year={2025}
}

@article{ma2025follow,
  title={Follow-your-motion: Video motion transfer via efficient spatial-temporal decoupled finetuning},
  author={Ma, Yue and Liu, Yulong and Zhu, Qiyuan and Yang, Ayden and Feng, Kunyu and Zhang, Xinhua and Yan, Zexuan and Li, Zhifeng and Han, Sirui and Qi, Chenyang and others},
  journal={arXiv preprint arXiv:2506.05207},
  year={2025}
}

@misc{relax_forcing,
    title={Relax Forcing: Relaxed KV-Memory for Consistent Long Video Generation},
    author={Zengqun Zhao and Yanzuo Lu and Ziquan Liu and Jifei Song and Jiankang Deng and Ioannis Patras},
    year={2026},
    eprint={2603.21366},
    archivePrefix={arXiv},
    primaryClass={cs.CV},
    url={https://arxiv.org/abs/2603.21366}
}

@misc{monarch_rt,
    title={MonarchRT: Efficient Attention for Real-Time Video Generation},
    author={Krish Agarwal and Zhuoming Chen and Cheng Luo and Yongqi Chen and Haizhong Zheng and Xun Huang and Atri Rudra and Beidi Chen},
    year={2026},
    eprint={2602.12271},
    archivePrefix={arXiv},
    primaryClass={cs.CV},
    url={https://arxiv.org/abs/2602.12271}
}

@inproceedings{self_forcing++,
    title={Self-Forcing++: Towards Minute-Scale High-Quality Video Generation},
    author={Justin Cui and Jie Wu and Ming Li and Tao Yang and Xiaojie Li and Rui Wang and Andrew Bai and Yuanhao Ban and Cho-Jui Hsieh},
    booktitle={The Fourteenth International Conference on Learning Representations},
    year={2026},
    url={https://openreview.net/forum?id=DzvPiqh23f}
}

@article{vbench++,
    author={Huang, Ziqi and Zhang, Fan and Xu, Xiaojie and He, Yinan and Yu, Jiashuo and Dong, Ziyue and Ma, Qianli and Chanpaisit, Nattapol and Si, Chenyang and Jiang, Yuming and Wang, Yaohui and Chen, Xinyuan and Chen, Ying-Cong and Wang, Limin and Lin, Dahua and Qiao, Yu and Liu, Ziwei},
    journal={IEEE Transactions on Pattern Analysis and Machine Intelligence}, 
    title={VBench++: Comprehensive and Versatile Benchmark Suite for Video Generative Models}, 
    year={2026},
    volume={48},
    number={3},
    pages={3268-3285},
    doi={10.1109/TPAMI.2025.3633890}
}

@misc{moviegen,
    title={Movie Gen: A Cast of Media Foundation Models}, 
    author={Adam Polyak and Amit Zohar and Andrew Brown and Andros Tjandra and Animesh Sinha and Ann Lee and Apoorv Vyas and Bowen Shi and Chih-Yao Ma and Ching-Yao Chuang and David Yan and Dhruv Choudhary and Dingkang Wang and Geet Sethi and Guan Pang and Haoyu Ma and Ishan Misra and Ji Hou and Jialiang Wang and Kiran Jagadeesh and Kunpeng Li and Luxin Zhang and Mannat Singh and Mary Williamson and Matt Le and Matthew Yu and Mitesh Kumar Singh and Peizhao Zhang and Peter Vajda and Quentin Duval and Rohit Girdhar and Roshan Sumbaly and Sai Saketh Rambhatla and Sam Tsai and Samaneh Azadi and Samyak Datta and Sanyuan Chen and Sean Bell and Sharadh Ramaswamy and Shelly Sheynin and Siddharth Bhattacharya and Simran Motwani and Tao Xu and Tianhe Li and Tingbo Hou and Wei-Ning Hsu and Xi Yin and Xiaoliang Dai and Yaniv Taigman and Yaqiao Luo and Yen-Cheng Liu and Yi-Chiao Wu and Yue Zhao and Yuval Kirstain and Zecheng He and Zijian He and Albert Pumarola and Ali Thabet and Artsiom Sanakoyeu and Arun Mallya and Baishan Guo and Boris Araya and Breena Kerr and Carleigh Wood and Ce Liu and Cen Peng and Dimitry Vengertsev and Edgar Schonfeld and Elliot Blanchard and Felix Juefei-Xu and Fraylie Nord and Jeff Liang and John Hoffman and Jonas Kohler and Kaolin Fire and Karthik Sivakumar and Lawrence Chen and Licheng Yu and Luya Gao and Markos Georgopoulos and Rashel Moritz and Sara K. Sampson and Shikai Li and Simone Parmeggiani and Steve Fine and Tara Fowler and Vladan Petrovic and Yuming Du},
    year={2025},
    eprint={2410.13720},
    archivePrefix={arXiv},
    primaryClass={cs.CV},
    url={https://arxiv.org/abs/2410.13720}, 
}

@misc{qwen3,
    title={Qwen3 Technical Report}, 
    author={An Yang and Anfeng Li and Baosong Yang and Beichen Zhang and Binyuan Hui and Bo Zheng and Bowen Yu and Chang Gao and Chengen Huang and Chenxu Lv and Chujie Zheng and Dayiheng Liu and Fan Zhou and Fei Huang and Feng Hu and Hao Ge and Haoran Wei and Huan Lin and Jialong Tang and Jian Yang and Jianhong Tu and Jianwei Zhang and Jianxin Yang and Jiaxi Yang and Jing Zhou and Jingren Zhou and Junyang Lin and Kai Dang and Keqin Bao and Kexin Yang and Le Yu and Lianghao Deng and Mei Li and Mingfeng Xue and Mingze Li and Pei Zhang and Peng Wang and Qin Zhu and Rui Men and Ruize Gao and Shixuan Liu and Shuang Luo and Tianhao Li and Tianyi Tang and Wenbiao Yin and Xingzhang Ren and Xinyu Wang and Xinyu Zhang and Xuancheng Ren and Yang Fan and Yang Su and Yichang Zhang and Yinger Zhang and Yu Wan and Yuqiong Liu and Zekun Wang and Zeyu Cui and Zhenru Zhang and Zhipeng Zhou and Zihan Qiu},
    year={2025},
    eprint={2505.09388},
    archivePrefix={arXiv},
    primaryClass={cs.CL},
    url={https://arxiv.org/abs/2505.09388}, 
}

@inproceedings{deepcache,
    title={DeepCache: Accelerating Diffusion Models for Free},
    author={Ma, Xinyin and Fang, Gongfan and Wang, Xinchao},
    booktitle={The IEEE/CVF Conference on Computer Vision and Pattern Recognition},
    year={2024}
}

@misc{fora,
    title={FORA: Fast-Forward Caching in Diffusion Transformer Acceleration}, 
    author={Pratheba Selvaraju and Tianyu Ding and Tianyi Chen and Ilya Zharkov and Luming Liang},
    year={2024},
    eprint={2407.01425},
    archivePrefix={arXiv},
    primaryClass={cs.CV},
    url={https://arxiv.org/abs/2407.01425}, 
}

@article{teacache,
    title={Timestep Embedding Tells: It's Time to Cache for Video Diffusion Model},
    author={Liu, Feng and Zhang, Shiwei and Wang, Xiaofeng and Wei, Yujie and Qiu, Haonan and Zhao, Yuzhong and Zhang, Yingya and Ye, Qixiang and Wan, Fang},
    journal={arXiv preprint arXiv:2411.19108},
    year={2024}
}

@inproceedings{toca,
    title={Accelerating Diffusion Transformers with Token-wise Feature Caching},
    author={Chang Zou and Xuyang Liu and Ting Liu and Siteng Huang and Linfeng Zhang},
    booktitle={The Thirteenth International Conference on Learning Representations},
    year={2025},
    url={https://openreview.net/forum?id=yYZbZGo4ei}
}

@inproceedings{taylorseer,
    author    = {Liu, Jiacheng and Zou, Chang and Lyu, Yuanhuiyi and Chen, Junjie and Zhang, Linfeng},
    title     = {From Reusing to Forecasting: Accelerating Diffusion Models with TaylorSeers},
    booktitle = {Proceedings of the IEEE/CVF International Conference on Computer Vision (ICCV)},
    month     = {October},
    year      = {2025},
    pages     = {15853-15863}
}

@misc{lesa,
    title={LESA: Learnable Stage-Aware Predictors for Diffusion Model Acceleration}, 
    author={Peiliang Cai and Jiacheng Liu and Haowen Xu and Xinyu Wang and Chang Zou and Linfeng Zhang},
    year={2026},
    eprint={2602.20497},
    archivePrefix={arXiv},
    primaryClass={cs.CV},
    url={https://arxiv.org/abs/2602.20497}, 
}

@inproceedings{streamingllm,
    author = {Xiao, Guangxuan and Tian, Yuandong and Chen, Beidi and Han, Song and Lewis, Mike },
    booktitle = {International Conference on Learning Representations},
    editor = {B. Kim and Y. Yue and S. Chaudhuri and K. Fragkiadaki and M. Khan and Y. Sun},
    pages = {21875--21895},
    title = {Efficient Streaming Language Models with Attention Sinks},
    url = {https://proceedings.iclr.cc/paper_files/paper/2024/file/5e5fd18f863cbe6d8ae392a93fd271c9-Paper-Conference.pdf},
    volume = {2024},
    year = {2024}
}

@inproceedings{h2o,
    author = {Zhang, Zhenyu and Sheng, Ying and Zhou, Tianyi and Chen, Tianlong and Zheng, Lianmin and Cai, Ruisi and Song, Zhao and Tian, Yuandong and R\'{e}, Christopher and Barrett, Clark and Wang, Zhangyang "Atlas" and Chen, Beidi},
    booktitle = {Advances in Neural Information Processing Systems},
    editor = {A. Oh and T. Naumann and A. Globerson and K. Saenko and M. Hardt and S. Levine},
    pages = {34661--34710},
    publisher = {Curran Associates, Inc.},
    title = {H2O: Heavy-Hitter Oracle for Efficient Generative Inference of Large Language Models},
    url = {https://proceedings.neurips.cc/paper_files/paper/2023/file/6ceefa7b15572587b78ecfcebb2827f8-Paper-Conference.pdf},
    volume = {36},
    year = {2023}
}

@inproceedings{duoattention,
    title={DuoAttention: Efficient Long-Context {LLM} Inference with Retrieval and Streaming Heads},
    author={Guangxuan Xiao and Jiaming Tang and Jingwei Zuo and junxian guo and Shang Yang and Haotian Tang and Yao Fu and Song Han},
    booktitle={The Thirteenth International Conference on Learning Representations},
    year={2025},
    url={https://openreview.net/forum?id=cFu7ze7xUm}
}

@misc{fastgen,
    title={DeepSpeed-FastGen: High-throughput Text Generation for LLMs via MII and DeepSpeed-Inference}, 
    author={Connor Holmes and Masahiro Tanaka and Michael Wyatt and Ammar Ahmad Awan and Jeff Rasley and Samyam Rajbhandari and Reza Yazdani Aminabadi and Heyang Qin and Arash Bakhtiari and Lev Kurilenko and Yuxiong He},
    year={2024},
    eprint={2401.08671},
    archivePrefix={arXiv},
    primaryClass={cs.PF},
    url={https://arxiv.org/abs/2401.08671}, 
}

@misc{hunyuanvideo2025,
    title={HunyuanVideo 1.5 Technical Report}, 
    author={Bing Wu and Chang Zou and Changlin Li and Duojun Huang and Fang Yang and Hao Tan and Jack Peng and Jianbing Wu and Jiangfeng Xiong and Jie Jiang and Linus and Patrol and Peizhen Zhang and Peng Chen and Penghao Zhao and Qi Tian and Songtao Liu and Weijie Kong and Weiyan Wang and Xiao He and Xin Li and Xinchi Deng and Xuefei Zhe and Yang Li and Yanxin Long and Yuanbo Peng and Yue Wu and Yuhong Liu and Zhenyu Wang and Zuozhuo Dai and Bo Peng and Coopers Li and Gu Gong and Guojian Xiao and Jiahe Tian and Jiaxin Lin and Jie Liu and Jihong Zhang and Jiesong Lian and Kaihang Pan and Lei Wang and Lin Niu and Mingtao Chen and Mingyang Chen and Mingzhe Zheng and Miles Yang and Qiangqiang Hu and Qi Yang and Qiuyong Xiao and Runzhou Wu and Ryan Xu and Rui Yuan and Shanshan Sang and Shisheng Huang and Siruis Gong and Shuo Huang and Weiting Guo and Xiang Yuan and Xiaojia Chen and Xiawei Hu and Wenzhi Sun and Xiele Wu and Xianshun Ren and Xiaoyan Yuan and Xiaoyue Mi and Yepeng Zhang and Yifu Sun and Yiting Lu and Yitong Li and You Huang and Yu Tang and Yixuan Li and Yuhang Deng and Yuan Zhou and Zhichao Hu and Zhiguang Liu and Zhihe Yang and Zilin Yang and Zhenzhi Lu and Zixiang Zhou and Zhao Zhong},
    year={2025},
    eprint={2511.18870},
    archivePrefix={arXiv},
    primaryClass={cs.CV},
    url={https://arxiv.org/abs/2511.18870}, 
}

@misc{ltx2,
    title={LTX-2: Efficient Joint Audio-Visual Foundation Model}, 
    author={Yoav HaCohen and Benny Brazowski and Nisan Chiprut and Yaki Bitterman and Andrew Kvochko and Avishai Berkowitz and Daniel Shalem and Daphna Lifschitz and Dudu Moshe and Eitan Porat and Eitan Richardson and Guy Shiran and Itay Chachy and Jonathan Chetboun and Michael Finkelson and Michael Kupchick and Nir Zabari and Nitzan Guetta and Noa Kotler and Ofir Bibi and Ori Gordon and Poriya Panet and Roi Benita and Shahar Armon and Victor Kulikov and Yaron Inger and Yonatan Shiftan and Zeev Melumian and Zeev Farbman},
    year={2026},
    eprint={2601.03233},
    archivePrefix={arXiv},
    primaryClass={cs.CV},
    url={https://arxiv.org/abs/2601.03233}, 
}

@misc{seedance2,
    title={Seedance 2.0: Advancing Video Generation for World Complexity}, 
    author={Team Seedance and De Chen and Liyang Chen and Xin Chen and Ying Chen and Zhuo Chen and Zhuowei Chen and Feng Cheng and Tianheng Cheng and Yufeng Cheng and Mojie Chi and Xuyan Chi and Jian Cong and Qinpeng Cui and Fei Ding and Qide Dong and Yujiao Du and Haojie Duanmu and Junliang Fan and Jiarui Fang and Jing Fang and Zetao Fang and Chengjian Feng and Yu Gao and Diandian Gu and Dong Guo and Hanzhong Guo and Qiushan Guo and Boyang Hao and Hongxiang Hao and Haoxun He and Jiaao He and Qian He and Tuyen Hoang and Heng Hu and Ruoqing Hu and Yuxiang Hu and Jiancheng Huang and Weilin Huang and Zhaoyang Huang and Zhongyi Huang and Jishuo Jin and Ming Jing and Ashley Kim and Shanshan Lao and Yichong Leng and Bingchuan Li and Gen Li and Haifeng Li and Huixia Li and Jiashi Li and Ming Li and Xiaojie Li and Xingxing Li and Yameng Li and Yiying Li and Yu Li and Yueyan Li and Chao Liang and Han Liang and Jianzhong Liang and Ying Liang and Wang Liao and J. H. Lien and Shanchuan Lin and Xi Lin and Feng Ling and Yue Ling and Fangfang Liu and Jiawei Liu and Jihao Liu and Jingtuo Liu and Shu Liu and Sichao Liu and Wei Liu and Xue Liu and Zuxi Liu and Ruijie Lu and Lecheng Lyu and Jingting Ma and Tianxiang Ma and Xiaonan Nie and Jingzhe Ning and Junjie Pan and Xitong Pan and Ronggui Peng and Xueqiong Qu and Yuxi Ren and Yuchen Shen and Guang Shi and Lei Shi and Yinglong Song and Fan Sun and Li Sun and Renfei Sun and Wenjing Tang and Boyang Tao and Zirui Tao and Dongliang Wang and Feng Wang and Hulin Wang and Ke Wang and Qingyi Wang and Rui Wang and Shuai Wang and Shulei Wang and Weichen Wang and Xuanda Wang and Yanhui Wang and Yue Wang and Yuping Wang and Yuxuan Wang and Zijie Wang and Ziyu Wang and Guoqiang Wei and Meng Wei and Di Wu and Guohong Wu and Hanjie Wu and Huachao Wu and Jian Wu and Jie Wu and Ruolan Wu and Shaojin Wu and Xiaohu Wu and Xinglong Wu and Yonghui Wu and Ruiqi Xia and Xin Xia and Xuefeng Xiao and Shuang Xu and Bangbang Yang and Jiaqi Yang and Runkai Yang and Tao Yang and Yihang Yang and Zhixian Yang and Ziyan Yang and Fulong Ye and Bingqian Yi and Xing Yin and Yongbin You and Linxiao Yuan and Weihong Zeng and Xuejiao Zeng and Yan Zeng and Siyu Zhai and Zhonghua Zhai and Bowen Zhang and Chenlin Zhang and Heng Zhang and Jun Zhang and Manlin Zhang and Peiyuan Zhang and Shuo Zhang and Xiaohe Zhang and Xiaoying Zhang and Xinyan Zhang and Xinyi Zhang and Yichi Zhang and Zixiang Zhang and Haiyu Zhao and Huating Zhao and Liming Zhao and Yian Zhao and Guangcong Zheng and Jianbin Zheng and Xiaozheng Zheng and Zerong Zheng and Kuan Zhu and Feilong Zuo},
    year={2026},
    eprint={2604.14148},
    archivePrefix={arXiv},
    primaryClass={cs.CV},
    url={https://arxiv.org/abs/2604.14148}, 
}

@article{lingbot-world,
    title={Advancing Open-source World Models}, 
    author={Robbyant Team and Zelin Gao and Qiuyu Wang and Yanhong Zeng and Jiapeng Zhu and Ka Leong Cheng and Yixuan Li and Hanlin Wang and Yinghao Xu and Shuailei Ma and Yihang Chen and Jie Liu and Yansong Cheng and Yao Yao and Jiayi Zhu and Yihao Meng and Kecheng Zheng and Qingyan Bai and Jingye Chen and Zehong Shen and Yue Yu and Xing Zhu and Yujun Shen and Hao Ouyang},
    journal={arXiv preprint arXiv:2601.20540},
    year={2026}
}

@misc{Hunyuan-GameCraft-2,
    title={Hunyuan-GameCraft-2: Instruction-following Interactive Game World Model}, 
    author={Junshu Tang and Jiacheng Liu and Jiaqi Li and Longhuang Wu and Haoyu Yang and Penghao Zhao and Siruis Gong and Xiang Yuan and Shuai Shao and Linfeng Zhang and Qinglin Lu},
    year={2026},
    eprint={2511.23429},
    archivePrefix={arXiv},
    primaryClass={cs.CV},
    url={https://arxiv.org/abs/2511.23429}, 
}

@article{helios,
    title={Helios: Real Real-Time Long Video Generation Model},
    author={Yuan, Shenghai and Yin, Yuanyang and Li, Zongjian and Huang, Xinwei and Yang, Xiao and Yuan, Li},
    journal={arXiv preprint arXiv:2603.04379},
    year={2026}
}

@misc{ddpm,
    title={Denoising Diffusion Probabilistic Models}, 
    author={Jonathan Ho and Ajay Jain and Pieter Abbeel},
    year={2020},
    eprint={2006.11239},
    archivePrefix={arXiv},
    primaryClass={cs.LG},
    url={https://arxiv.org/abs/2006.11239}, 
}

@inproceedings{ddim,
    title     = {Denoising Diffusion Implicit Models},
    author    = {Jiaming Song and Chenlin Meng and Stefano Ermon},
    booktitle = {International Conference on Learning Representations (ICLR)},
    year      = {2021}
}

@article{blattmann2023SVD,
    title        = {Stable video diffusion: Scaling latent video diffusion models to large datasets},
    author       = {Blattmann, Andreas and Dockhorn, Tim and Kulal, Sumith and Mendelevitch, Daniel and Kilian, Maciej and Lorenz, Dominik and Levi, Yam and English, Zion and Voleti, Vikram and Letts, Adam and others},
    year         = 2023,
    journal      = {arXiv preprint arXiv:2311.15127}
}

@misc{dit,
    title={Scalable Diffusion Models with Transformers}, 
    author={William Peebles and Saining Xie},
    year={2023},
    eprint={2212.09748},
    archivePrefix={arXiv},
    primaryClass={cs.CV},
    url={https://arxiv.org/abs/2212.09748}, 
}

@article{lu2022dpmsolverpp,
   title={DPM-Solver++: Fast Solver for Guided Sampling of Diffusion Probabilistic Models},
   volume={22},
   ISSN={2731-5398},
   url={http://dx.doi.org/10.1007/s11633-025-1562-4},
   DOI={10.1007/s11633-025-1562-4},
   number={4},
   journal={Machine Intelligence Research},
   publisher={Springer Science and Business Media LLC},
   author={Lu, Cheng and Zhou, Yuhao and Bao, Fan and Chen, Jianfei and Li, Chongxuan and Zhu, Jun},
   year={2025},
   month=jun, pages={730–751} 
}

@inproceedings{salimans2022progressive,
    title     = {Progressive Distillation for Fast Sampling of Diffusion Models},
    author    = {Tim Salimans and Jonathan Ho},
    booktitle = {International Conference on Learning Representations (ICLR)},
    year      = {2022}
}

@article{luo2023latentconsistencymodelssynthesizing,
    title={Latent consistency models: Synthesizing high-resolution images with few-step inference},
    author={Luo, Simian and Tan, Yiqin and Huang, Longbo and Li, Jian and Zhao, Hang},
    journal={arXiv preprint arXiv:2310.04378},
    year={2023}
}

@inproceedings{zhao2023unipc,
    title     = {UniPC: A Unified Predictor-Corrector Framework for Fast Sampling of Diffusion Models},
    author={Wenliang Zhao and Lujia Bai and Yongming Rao and Jie Zhou and Jiwen Lu},
    booktitle = {Advances in Neural Information Processing Systems (NeurIPS)},
    year      = {2023}
}

@misc{geng2025meanflowsonestepgenerative,
    title={Mean Flows for One-step Generative Modeling}, 
    author={Zhengyang Geng and Mingyang Deng and Xingjian Bai and J. Zico Kolter and Kaiming He},
    year={2025},
    eprint={2505.13447},
    archivePrefix={arXiv},
    primaryClass={cs.LG},
    url={https://arxiv.org/abs/2505.13447}, 
}

@misc{frans2025stepdiffusionshortcutmodels,
    title={One Step Diffusion via Shortcut Models}, 
    author={Kevin Frans and Danijar Hafner and Sergey Levine and Pieter Abbeel},
    year={2025},
    eprint={2410.12557},
    archivePrefix={arXiv},
    primaryClass={cs.LG},
    url={https://arxiv.org/abs/2410.12557}, 
}

@misc{SageAttention,
    title={SageAttention: Accurate 8-Bit Attention for Plug-and-play Inference Acceleration}, 
    author={Jintao Zhang and Jia Wei and Haofeng Huang and Pengle Zhang and Jun Zhu and Jianfei Chen},
    year={2025},
    eprint={2410.02367},
    archivePrefix={arXiv},
    primaryClass={cs.LG},
    url={https://arxiv.org/abs/2410.02367}, 
}

@misc{SageAttention2,
    title={SageAttention2: Efficient Attention with Thorough Outlier Smoothing and Per-thread INT4 Quantization}, 
    author={Jintao Zhang and Haofeng Huang and Pengle Zhang and Jia Wei and Jun Zhu and Jianfei Chen},
    year={2025},
    eprint={2411.10958},
    archivePrefix={arXiv},
    primaryClass={cs.LG},
    url={https://arxiv.org/abs/2411.10958}, 
}

@inproceedings{li2024distrifusion,
    title={DistriFusion: Distributed Parallel Inference for High-Resolution Diffusion Models},
    author={Li, Muyang and Cai, Tianle and Cao, Jiaxin and Zhang, Qinsheng and Cai, Han and Bai, Junjie and Jia, Yangqing and Liu, Ming-Yu and Li, Kai and Han, Song},
    booktitle={Proceedings of the IEEE/CVF Conference on Computer Vision and Pattern Recognition (CVPR)},
    year={2024}
}

@inproceedings{fang2024pipefusion,
    title={PipeFusion: Patch-level Pipeline Parallelism for Diffusion Transformers Inference},
    author={Jiarui Fang and Jinzhe Pan and Aoyu Li and Xibo Sun and WANG Jiannan},
    booktitle={The Thirty-ninth Annual Conference on Neural Information Processing Systems (NeurIPS)},
    year={2025}
}

@article{ltx-video,
    title={Ltx-video: Realtime video latent diffusion},
    author={HaCohen, Yoav and Chiprut, Nisan and Brazowski, Benny and Shalem, Daniel and Moshe, Dudu and Richardson, Eitan and Levin, Eran and Shiran, Guy and Zabari, Nir and Gordon, Ori and others},
    journal={arXiv preprint arXiv:2501.00103},
    year={2024}
}

@article{cogvideox,
    title={Cogvideox: Text-to-video diffusion models with an expert transformer},
    author={Yang, Zhuoyi and Teng, Jiayan and Zheng, Wendi and Ding, Ming and Huang, Shiyu and Xu, Jiazheng and Yang, Yuanming and Hong, Wenyi and Zhang, Xiaohan and Feng, Guanyu and others},
    journal={arXiv preprint arXiv:2408.06072},
    year={2024}
}

@article{brooks2024video,
    title={Video generation models as world simulators},
    author={Brooks, Tim and Peebles, Bill and Holmes, Connor and DePue, Will and Guo, Yufei and Jing, Leo and Schnurr, David and Taylor, Joe and Luhman, Troy and Luhman, Eric and others},
    journal={OpenAI Blog},
    volume={1},
    number={8},
    pages={1},
    year={2024}
}

@article{diffusion_force,
    title={Diffusion forcing: Next-token prediction meets full-sequence diffusion},
    author={Chen, Boyuan and Mart{\'\i} Mons{\'o}, Diego and Du, Yilun and Simchowitz, Max and Tedrake, Russ and Sitzmann, Vincent},
    journal={Advances in Neural Information Processing Systems},
    volume={37},
    pages={24081--24125},
    year={2024}
}

@article{context_force,
    title={Context Forcing: Consistent Autoregressive Video Generation with Long Context},
    author={Chen, Shuo and Wei, Cong and Sun, Sun and Nie, Ping and Zhou, Kai and Zhang, Ge and Yang, Ming-Hsuan and Chen, Wenhu},
    journal={arXiv preprint arXiv:2602.06028},
    year={2026}
}

\newpage
\appendix

\section{Limitations}
\label{limitations}
Focused Forcing is designed as a training-free KV selection method for efficient autoregressive video diffusion. Experiments evaluate it under the VBench-Long protocol and several representative autoregressive video diffusion paradigms, where it demonstrates consistent speed-quality improvements. Broader evaluation on higher resolutions, other model architectures, and more diverse deployment settings would further validate its generality. Moreover, the current implementation adopts fixed KV-budget bounds and a fixed attention-diversity weight, which provide a favorable trade-off in our experiments but may be further adapted to different models, resolutions, or latency targets. The realized acceleration may also vary across hardware platforms and attention kernel implementations.

\section{Method Details}
\subsection{Estimation of Head Importance for Budget Allocation}
\label{app:head_importance}
\begin{algorithm}[H]
  \caption{Head Importance Estimation and KV-Budget Allocation}
  \small
  \begin{algorithmic}[1]
    \Require Prompt set $\mathcal{P}=\{p_m\}_{m=1}^{M}$;
             number of layers $L$;
             number of heads per layer $H$;
             rollout window length $K$;
             number of sampled windows $N_w$;
             minimum and maximum KV budgets $b_{\min}, b_{\max}$;
             budget curvature $\gamma$;
             fake-score model $S_{\mathrm{fake}}$;
             real-score model $S_{\mathrm{real}}$;
             CFG scale $s_{\mathrm{cfg}}$
    \Ensure Fixed per-head KV budget table $\Pi_{\mathrm{KV}}$
    
    \State Initialize head score table $\mathcal{S}\in\mathbb{R}^{L\times H}$ to zeros

    \For{$m=1,\dots,M$}
      \State Sample a random seed $s_m$ for prompt $p_m$
      \State Construct $B=L\times H$ rollouts, with rollout $b$ masking head $(\lfloor b/H\rfloor,\;b\bmod H)$
      \State Run denoising rollout and obtain masked latent trajectories $\{\mathbf{Z}^{(b)}\}_{b=0}^{B-1}$

      \For{$b=0$ to $B-1$}
        \State $\mathrm{DMsum}\gets 0$ \textcolor{green!50!black}{\Comment{Accumulates DM losses over sampled windows.}}
        \State Split $\mathbf{Z}^{(b)}$ into $N_w$ temporal windows $\{W_r^{(b)}\}_{r=0}^{N_w-1}$, each with length $K$

        \For{$r=0$ to $N_w-1$}
          \State Sample timestep $t$ and Gaussian noise $\epsilon$; add noise $W_{r,t}^{(b)}=\mathrm{AddNoise}(W_r^{(b)},\epsilon,t)$
          \State Predict fake clean window $\hat{W}^{\mathrm{fake}}_r=S_{\mathrm{fake}}(W_{r,t}^{(b)},p_m,t)$
          \State Predict real clean windows $\hat{W}^{\mathrm{real}}_{r,\mathrm{cond}}=S_{\mathrm{real}}(W_{r,t}^{(b)},p_m,t)$ and $\hat{W}^{\mathrm{real}}_{r,\mathrm{uncond}}=S_{\mathrm{real}}(W_{r,t}^{(b)},p_{\emptyset},t)$
          \State Apply CFG: $\hat{W}^{\mathrm{real}}_r=\hat{W}^{\mathrm{real}}_{r,\mathrm{cond}}+s_{\mathrm{cfg}}(\hat{W}^{\mathrm{real}}_{r,\mathrm{cond}}-\hat{W}^{\mathrm{real}}_{r,\mathrm{uncond}})$
          \State Compute DM gradient $g_r^{(b)}=\hat{W}^{\mathrm{fake}}_r-\hat{W}^{\mathrm{real}}_r$
          \State Normalize DM gradient $g_r^{(b)}\gets g_r^{(b)}/(|W_r^{(b)}-\hat{W}^{\mathrm{real}}_r|_{\mathrm{mean}}+\epsilon)$
          \State Compute DM loss $d_r^{(b)}=\frac{1}{2}\left\|W_r^{(b)}-\mathrm{sg}(W_r^{(b)}-g_r^{(b)})\right\|_2^2$
          \State $\mathrm{DMsum}\gets \mathrm{DMsum}+d_r^{(b)}$
        \EndFor

        \State $\mathcal{S}[\ell_b,h_b]\gets \mathcal{S}[\ell_b,h_b]+\mathrm{DMsum}/N_w$
      \EndFor
    \EndFor

    \State Average head scores over prompts: $\mathcal{S}\gets \mathcal{S}/M$ \textcolor{green!50!black}{\Comment{Higher score means stronger degradation.}}
    \State $S_{\min}\gets \min_{\ell,h}\mathcal{S}[\ell,h]$, $S_{\max}\gets \max_{\ell,h}\mathcal{S}[\ell,h]$

    \For{$\ell=0$ to $L-1$}
      \For{$h=0$ to $H-1$}
        \State Normalize head importance:
        $$\hat{S}_{\ell,h}
        =
        \frac{
            \mathcal{S}[\ell,h]-S_{\min}
        }{
            S_{\max}-S_{\min}+\epsilon
        }$$

        \State Map importance to KV budget:
        $$b_{\ell,h}
        =
        \mathrm{round}
        \left(
            b_{\min}
            +
            \hat{S}_{\ell,h}^{\gamma}
            \left(
                b_{\max}-b_{\min}
            \right)
        \right)$$
      \EndFor
    \EndFor

    \State Construct fixed per-head KV budget table $\Pi_{\mathrm{KV}}=\{b_{\ell,h}\}_{\ell=0,h=0}^{L-1,H-1}$ \textcolor{green!50!black}{\Comment{The budget table is computed offline.}}
    \State Freeze $\Pi_{\mathrm{KV}}$ for subsequent inference
  \end{algorithmic}
  \label{alg:head_prioritization}
\end{algorithm}

\paragraph{Offline Profiling Cost.}
We use 32 calibration prompts and evaluate 360 layer-head pairs through 10-second autoregressive rollouts. This is a one-time offline cost and is excluded from online inference latency. The resulting budget table can be reused as long as the model weights and attention architecture remain unchanged.

\subsection{Proof of the Temporal Effect of Causal RoPE}
\label{app:temporal_rope}
We show that causal RoPE makes the attention logit explicitly depend on the relative temporal distance between a query frame and a key frame. This explains why the same historical frame can receive different attention scores as its relative temporal distance to the current frame changes.

Let $d=2m$ be the dimension of one attention head. RoPE groups a feature vector into $m$ two-dimensional blocks. For the $j$-th block, define the rotation matrix
$$
R(\theta)
:=
\left[
\begin{array}{cc}
\cos\theta & -\sin\theta \\
\sin\theta & \cos\theta
\end{array}
\right].
$$
In causal RoPE, a token at frame index $t$ is assigned a temporal rotary phase. Let $\mathcal T$ denote the index set of two-dimensional rotary blocks associated with temporal encoding, and let $\omega_j$ be the temporal frequency for $j\in\mathcal T$. For a token at frame index $t$, the $j$-th temporal block is rotated by angle $\omega_j t$.

For a query token at frame $t_q$ and a key token at frame $t_k$, denote their pre-RoPE representations in the $j$-th temporal block as
$$
q_j :=
\left[
\begin{array}{c}
q_{2j-1} \\
q_{2j}
\end{array}
\right],
\qquad
k_j :=
\left[
\begin{array}{c}
k_{2j-1} \\
k_{2j}
\end{array}
\right].
$$
After applying temporal RoPE, we have
$$
\widetilde q_j = R(\omega_j t_q)q_j,
\qquad
\widetilde k_j = R(\omega_j t_k)k_j.
$$
\paragraph{Proposition.}
The temporal RoPE contribution to the attention logit is an explicit function of the relative temporal distance $\Delta t=t_k-t_q$. In particular,
$$
\ell_T(\Delta t)
=
\frac{1}{\sqrt d}
\sum_{j\in\mathcal T}
\left[
A_j\cos(\omega_j\Delta t)
+
B_j\sin(\omega_j\Delta t)
\right],
$$
where
$$
A_j=q_{2j-1}k_{2j-1}+q_{2j}k_{2j},
\qquad
B_j=q_{2j}k_{2j-1}-q_{2j-1}k_{2j}.
$$
\paragraph{Proof.}
The contribution of the $j$-th temporal block to the attention logit is
$$
\widetilde q_j^\top \widetilde k_j
=
\left(R(\omega_j t_q)q_j\right)^\top
\left(R(\omega_j t_k)k_j\right).
$$
Using the identity $R(a)^\top R(b)=R(b-a)$, we obtain
$$
\begin{array}{rcl}
\widetilde q_j^\top \widetilde k_j
=
q_j^\top R(\omega_j t_q)^\top R(\omega_j t_k)k_j
=
q_j^\top R(\omega_j(t_k-t_q))k_j.
\end{array}
$$
Let $\Delta t=t_k-t_q$. Then
$$
\widetilde q_j^\top \widetilde k_j
=
q_j^\top R(\omega_j\Delta t)k_j.
$$
Expanding the rotation matrix gives
$$
\begin{array}{rcl}
q_j^\top R(\omega_j\Delta t)k_j
&=&
\left[
\begin{array}{cc}
q_{2j-1} & q_{2j}
\end{array}
\right]
\left[
\begin{array}{cc}
\cos(\omega_j\Delta t) & -\sin(\omega_j\Delta t) \\
\sin(\omega_j\Delta t) & \cos(\omega_j\Delta t)
\end{array}
\right]
\left[
\begin{array}{c}
k_{2j-1} \\
k_{2j}
\end{array}
\right]
\\[3mm]
&=&
(q_{2j-1}k_{2j-1}+q_{2j}k_{2j})
\cos(\omega_j\Delta t)
+
(q_{2j}k_{2j-1}-q_{2j-1}k_{2j})
\sin(\omega_j\Delta t).
\end{array}
$$
Substituting the definitions of $A_j$ and $B_j$, we have
$$
q_j^\top R(\omega_j\Delta t)k_j
=
A_j\cos(\omega_j\Delta t)
+
B_j\sin(\omega_j\Delta t).
$$
Summing over all temporal rotary blocks and applying the standard attention scaling factor $1/\sqrt d$, we obtain
$$
\ell_T(\Delta t)
=
\frac{1}{\sqrt d}
\sum_{j\in\mathcal T}
\left[
A_j\cos(\omega_j\Delta t)
+
B_j\sin(\omega_j\Delta t)
\right].
$$
Therefore, the temporal part of the attention logit explicitly depends on the relative temporal distance $\Delta t$. This proves the proposition.
\paragraph{General case.}
In the non-degenerate case, there exists at least one temporal block $j\in\mathcal T$ such that $q_j\neq 0$ and $k_j\neq 0$. For this block,
$$
A_j^2+B_j^2
=
\left(q_{2j-1}^2+q_{2j}^2\right)
\left(k_{2j-1}^2+k_{2j}^2\right)
=
\|q_j\|^2\|k_j\|^2
>
0.
$$
Thus, at least one of $A_j$ and $B_j$ is nonzero. Therefore,
$$
A_j\cos(\omega_j\Delta t)
+
B_j\sin(\omega_j\Delta t)
$$
is not a constant function of $\Delta t$. Hence, in general, changing the relative temporal distance changes the attention logit. Since attention weights are computed by a softmax, such logit changes may be further amplified in the final attention distribution.

\paragraph{Special cases.}
The temporal effect may disappear or become weak in the following degenerate cases.

\noindent{\it Case 1: Same temporal position.}
If $t_q=t_k$, then $\Delta t=0$ and $R(\omega_j\Delta t)=I$. Hence,
$$
q_j^\top R(\omega_j\Delta t)k_j
=
q_j^\top k_j.
$$
In this case, temporal RoPE does not alter the inner product.

\noindent{\it Case 2: Zero temporal components.}
If $q_j=0$ or $k_j=0$ for all temporal blocks $j\in\mathcal T$, then all temporal coefficients vanish, namely $A_j=B_j=0$. Hence,
$$
\ell_T(\Delta t)=0.
$$
In this case, temporal RoPE has no effect on the attention logit.

\noindent{\it Case 3: Periodic phase alignment.}
If two temporal distances $\Delta t_1$ and $\Delta t_2$ satisfy
$$
\omega_j(\Delta t_2-\Delta t_1)
\in
\{2\pi z: z=0,\pm1,\pm2,\ldots\}
$$
for all active temporal frequencies $j$, then the temporal phases are identical. Hence,
$$
\ell_T(\Delta t_1)=\ell_T(\Delta t_2).
$$
\noindent{\it Case 4: Frequency cancellation.}
Different frequency components may partially cancel each other at particular values of $\Delta t$, reducing the overall temporal effect.

\subsection{Computational Complexity and Extra Memory Overhead}
We analyze the computation and extra memory introduced by packed variable-length FlashAttention. In each attention call, the current chunk contains $QF=3$ query frames. The model has $L=30$ layers and $H=12$ heads per layer. For each layer-head pair $(\ell,h)$, Focused Forcing assigns a KV budget $b_{\ell,h}$, which denotes the number of historical key-value frames retained per query frame. With $b_{\min}=4$, $b_{\max}=12$, and $\gamma=2$, the allocated budgets satisfy
$$
\sum_{\ell=0}^{29}\sum_{h=0}^{11} b_{\ell,h}=1958.
$$
Thus, the packed frame-level attention cost is
$$
C_{\mathrm{pack}}
=
QF\sum_{\ell,h} b_{\ell,h}
=
3\times 1958
=
5874.
$$
For dense attention with a $21$-frame KV window,
$$
C_{\mathrm{dense}}
=
L\times H\times QF\times F_{\mathrm{dense}}
=
30\times 12\times 3\times 21
=
22680.
$$
Therefore, packed attention uses
$$
\frac{C_{\mathrm{pack}}}{C_{\mathrm{dense}}}
=
\frac{5874}{22680}
=
0.259.
$$
of the dense frame-level attention computation, corresponding to a $74.1\%$ reduction and a theoretical frame-level speedup of $3.86\times$.

We next estimate the extra QKV packing memory. Each frame contains $N=1560$ tokens. The per-head dimension is $D=128$, and BF16 uses $s=2$ bytes per element. Thus, one frame in one head occupies
$$
N\times D\times s
=
1560\times 128\times 2
=
399360\ \mathrm{bytes}
=
0.381\ \mathrm{MiB}.
$$
The packed query tensor contains all $QF=3$ query frames and all $H=12$ heads:
$$
M_Q
=
QF\times H\times N\times D\times s
=
13.71\ \mathrm{MiB}.
$$
For layer $\ell$, let
$
S_\ell=\sum_{h=0}^{11} b_{\ell,h},
$
the packed key and value tensors together require
$$
M_{K+V}^{(\ell)}
=
2\times QF\times S_\ell\times N\times D\times s.
$$
Hence the total extra packing memory in layer $\ell$ is
$$
M_{\mathrm{pack}}^{(\ell)}
=
M_Q+M_{K+V}^{(\ell)}.
$$
From the allocated budgets, the layer-wise sums satisfy
$$
S_{\min}=61,\qquad
S_{\mathrm{avg}}=\frac{1958}{30}=65.27,\qquad
S_{\max}=72.
$$
Substituting these values gives
$$
M_{\mathrm{pack}}^{\min}=153.10\ \mathrm{MiB},
\qquad
M_{\mathrm{pack}}^{\mathrm{avg}}=162.86\ \mathrm{MiB},
\qquad
M_{\mathrm{pack}}^{\max}=178.24\ \mathrm{MiB}.
$$
Therefore, when only $Q_{\mathrm{pack}}$, $K_{\mathrm{pack}}$, and $V_{\mathrm{pack}}$ are counted, the extra packing memory per attention call is
$$
153.10\ \mathrm{MiB}
\le
M_{\mathrm{pack}}^{(\ell)}
\le
178.24\ \mathrm{MiB},
$$
with an average of $162.86\ \mathrm{MiB}$. This overhead is layer-local because the packed tensors are temporary buffers used only during the current attention computation.

\section{Experimental Results and Details}
\subsection{Implementation Details}
\label{app:implementation}

For each prompt, all methods generate an approximately 30s video with 486 frames at 16 FPS and a resolution of $832 \times 480$, except that NOVA is evaluated at $768 \times 480$ due to its fixed training shape. All experiments are conducted on 8 NVIDIA A100 GPUs. For Focused Forcing, unless otherwise specified, we set the KV budget to $(b_{\min}, b_{\max}) = (4, 12)$ and the attention weight to $0.5$.

\subsection{Evaluation Metrics}
\label{app:metrics}

For generation quality, we report six visual metrics, including Subject Consistency, Background Consistency, Motion Smoothness, Dynamic Degree, Aesthetic Quality, and Imaging Quality. We further report aggregated scores, where Temporal Quality averages the first four metrics, Frame-wise Quality averages the last two, and Visual Quality averages all six. Text Alignment is computed by averaging Overall Consistency and CLIP Score.

\subsection{Detailed Ablation Results}
\label{app:ablation}

\paragraph{Ablation on KV budget.}
The KV budget defines the head-wise capacity for retaining historical KV entries and therefore controls the trade-off between temporal-context preservation and inference efficiency, as reported in Tab.~\ref{tab:kv_budget_ablation}. Smaller budgets yield higher attention speedup by aggressively pruning historical KV entries, but they may discard useful historical context and degrade visual quality. For example, $(4,6)$ achieves the highest attention speedup of $3.27\times$, but its visual quality drops to 74.67. In contrast, larger budgets preserve more temporal context and generally improve quality, while reducing the acceleration gain. Among the tested settings, $(b_{\min}, b_{\max})=(4,12)$ achieves the best visual quality of 80.00 with $2.77\times$ attention speedup and $1.45\times$ generation speedup, offering the most favorable balance between quality and efficiency.
\begin{table}[H]
    \vspace{-1.5mm}
    \centering
    \caption{Ablation experiments of kv budget on VBench-Long.}
    \vspace{-1.5mm}
    \resizebox{0.80\columnwidth}{!}{
        \setlength{\tabcolsep}{3pt}
        \renewcommand{\arraystretch}{1.1}
        \begin{tabular}{@{}l | c c c c | c c | c c@{}}
            \toprule
            \multirow{2}{*}{\makecell{KV Budget\\$(b_{\min}, b_{\max})$}} &
            \multirow{2}{*}{\makecell{Attn.\\Latency/s}} &
            \multirow{2}{*}{\makecell{Attn.\\Speedup}} &
            \multirow{2}{*}{\makecell{Gen.\\Latency/s}} &
            \multirow{2}{*}{\makecell{Gen.\\Speedup}} &
            \multirow{2}{*}{\makecell{Temporal\\Quality}} &
            \multirow{2}{*}{\makecell{Frame-wise\\Quality}} &
            \multirow{2}{*}{\makecell{Visual\\Quality}} &
            \multirow{2}{*}{\makecell{Text\\Alignment}} \\
            & & & & & & & & \\
            \midrule
            Self Forcing
            & 37.29 & 1.00$\times$ & 78.06 & 1.00$\times$ & 83.15 & 63.44 & 76.58 & 28.02 \\
            \midrule
            (4, 6)
            & 11.42 & 3.27$\times$ & 52.11 & 1.50$\times$ & 80.78 & 62.46 & 74.67 & 29.44 \\
            (4, 9)
            & 12.84 & 2.90$\times$ & 53.26 & 1.47$\times$ & 87.19 & 64.76 & 79.71 & 29.19 \\
            (4, 12)
            & 13.48 & 2.77$\times$ & 53.90 & 1.45$\times$ & 87.69 & 64.61 & 80.00 & 28.75 \\
            (4, 15)
            & 14.51 & 2.57$\times$ & 54.96 & 1.42$\times$ & 87.15 & 64.54 & 79.61 & 28.55 \\
            \midrule
            (5, 6)
            & 12.79 & 2.92$\times$ & 53.18 & 1.47$\times$ & 87.41 & 64.66 & 79.83 & 28.91 \\
            (5, 9)
            & 14.31 & 2.61$\times$ & 54.73 & 1.43$\times$ & 86.97 & 64.34 & 79.43 & 28.61 \\
            (5, 12)
            & 14.71 & 2.54$\times$ & 55.12 & 1.42$\times$ & 85.59 & 64.91 & 78.70 & 28.64 \\
            (5, 15)
            & 15.74 & 2.37$\times$ & 56.15 & 1.39$\times$ & 85.15 & 64.88 & 78.39 & 28.56 \\
            \midrule
            (6, 6)
            & 14.35 & 2.60$\times$ & 54.75 & 1.43$\times$ & 87.03 & 64.02 & 79.36 & 28.40 \\
            (6, 9)
            & 15.27 & 2.44$\times$ & 55.69 & 1.40$\times$ & 86.07 & 64.50 & 78.88 & 28.50 \\
            (6, 12)
            & 16.01 & 2.33$\times$ & 56.44 & 1.38$\times$ & 85.64 & 64.53 & 78.61 & 28.50 \\
            (6, 15)
            & 16.84 & 2.21$\times$ & 57.25 & 1.36$\times$ & 85.50 & 64.78 & 78.59 & 28.57 \\
            \midrule
            (7, 9)
            & 15.99 & 2.33$\times$ & 56.43 & 1.38$\times$ & 85.94 & 64.29 & 78.72 & 28.52 \\
            (7, 12)
            & 17.39 & 2.14$\times$ & 57.84 & 1.35$\times$ & 85.49 & 64.59 & 78.52 & 28.56 \\
            (7, 15)
            & 18.08 & 2.06$\times$ & 58.52 & 1.33$\times$ & 85.54 & 64.62 & 78.56 & 28.48 \\
            \midrule
            (8, 9)
            & 17.38 & 2.15$\times$ & 57.86 & 1.35$\times$ & 85.39 & 64.44 & 78.41 & 28.61 \\
            (8, 12)
            & 18.88 & 1.97$\times$ & 59.35 & 1.32$\times$ & 85.78 & 64.56 & 78.70 & 28.52 \\
            (8, 15)
            & 19.30 & 1.93$\times$ & 59.75 & 1.31$\times$ & 85.25 & 64.63 & 78.38 & 28.58 \\
            \bottomrule
        \end{tabular}
    }
    \label{tab:kv_budget_ablation}
    \vspace{-4mm}
\end{table}

The improvement in aggregate Visual Quality under the $(4,12)$ budget is largely driven by increased Dynamic Degree, rather than uniform improvements across all quality metrics. In contrast, the more conservative $(8,12)$ and $(8,15)$ budgets generally offer a better balance between inference efficiency and temporal consistency.

\vspace{-1.5mm}

\paragraph{Ablation on attention weight.}
The attention weight controls the relative importance of attention relevance and key diversity when selecting historical frames, with detailed results reported in Tab.~\ref{tab:attn_weight_ablation}. Compared with Self Forcing, all tested settings improve visual quality, indicating that our historical-frame selection consistently benefits generation quality. However, relying on either factor alone is suboptimal: using only diversity improves visual quality from 76.58 to 77.66 but remains clearly worse than balanced settings, while using only attention reaches 79.31 but can overemphasize redundant high-attention frames. Moderate attention weights achieve a better trade-off. In particular, a balanced setting of $0.5$ obtains the best visual quality of 80.00, while also maintaining strong text alignment. These results show that attention relevance and key diversity are complementary, and combining them is important for robust historical-frame selection.
\begin{table}[H]
    \vspace{-1.5mm}
    \centering
    \caption{Ablation experiments of attention weight on VBench-Long.}
    \vspace{-1.5mm}
    \resizebox{\columnwidth}{!}{
        \setlength{\tabcolsep}{3pt}
        \renewcommand{\arraystretch}{1.1}
        \begin{tabular}{@{}l | c c c c c c c c | c c@{}}
            \toprule
            \multirow{2}{*}{\makecell{Attention\\Weight}} &
            \multirow{2}{*}{\makecell{Subject\\Consistency}} &
            \multirow{2}{*}{\makecell{Background\\Consistency}} &
            \multirow{2}{*}{\makecell{Motion\\Smoothness}} &
            \multirow{2}{*}{\makecell{Dynamic\\Degree}} &
            \multirow{2}{*}{\makecell{Aesthetic\\Quality}} &
            \multirow{2}{*}{\makecell{Imaging\\Quality}} &
            \multirow{2}{*}{\makecell{Overall\\Consistency}} &
            \multirow{2}{*}{\makecell{Clip\\Score}} &
            \multirow{2}{*}{\makecell{Visual\\Quality}} &
            \multirow{2}{*}{\makecell{Text\\Alignment}} \\
            & & & & & & & & & & \\
            \midrule
            Self Forcing
            & 96.61 & 96.17 & 98.27 & 41.55 & 58.83 & 68.04 & 23.93 & 32.12 & 76.58 & 28.02 \\
            \midrule
            0.0
            & 94.12 & 94.92 & 97.86 & 51.95 & 59.76 & 67.36 & 24.77 & 33.05 & 77.66 & 28.91 \\
            0.1
            & 94.44 & 94.99 & 97.92 & 53.20 & 59.86 & 67.63 & 24.84 & 33.03 & 78.01 & 28.94 \\
            0.2
            & 95.00 & 95.08 & 97.96 & 56.82 & 59.99 & 67.99 & 24.67 & 32.97 & 78.81 & 28.82 \\
            0.3
            & 95.38 & 95.23 & 97.99 & 60.90 & 60.21 & 68.33 & 24.75 & 33.00 & 79.67 & 28.88 \\
            0.4
            & 95.69 & 95.34 & 97.94 & 61.75 & 60.10 & 68.34 & 24.45 & 32.93 & 79.86 & 28.69 \\
            0.5
            & 95.84 & 95.49 & 98.04 & 61.39 & 60.59 & 68.62 & 24.53 & 32.97 & 80.00 & 28.75 \\
            0.6
            & 96.12 & 95.66 & 98.16 & 57.23 & 60.52 & 68.58 & 24.53 & 33.05 & 79.38 & 28.79 \\
            0.7
            & 96.22 & 95.62 & 98.18 & 58.27 & 60.63 & 68.79 & 24.37 & 33.00 & 79.62 & 28.69 \\
            0.8
            & 96.19 & 95.68 & 98.19 & 56.68 & 60.64 & 68.65 & 24.29 & 32.97 & 79.34 & 28.63 \\
            0.9
            & 96.24 & 95.65 & 98.22 & 55.65 & 60.41 & 68.51 & 24.26 & 33.00 & 79.11 & 28.63 \\
            1.0
            & 96.32 & 95.64 & 98.23 & 56.75 & 60.55 & 68.40 & 24.40 & 33.09 & 79.31 & 28.75 \\
            \bottomrule
        \end{tabular}
    }
    \label{tab:attn_weight_ablation}
    \vspace{-4mm}
\end{table}

\paragraph{Ablation on Importance-Aware Budget Allocation.}
The importance-aware budget allocation controls how historical-frame budgets are distributed across attention heads, as summarized in Tab.~\ref{tab:budget_allocation_strategy_ablation}. Uniform budgets improve over Self Forcing when the budget is sufficiently large, e.g., $(8,8)$ and $(12,12)$ reach visual quality scores of 78.85 and 78.86, respectively, but they do not fully exploit the unequal importance of different heads. In contrast, the proposed normal allocation with $(4,12)$ achieves the best visual quality of 80.00, showing that assigning larger budgets to more important heads better preserves useful historical context. Reverse and random allocation under the same $(4,12)$ budget range perform worse, confirming that the gain comes from importance-aware allocation rather than merely using non-uniform budgets. These results validate the effectiveness of head-importance-guided KV budget allocation under compact cache budgets.
\begin{table}[H]
    \vspace{-1.5mm}
    \centering
    \caption{Ablation experiments of head-wise KV budget allocation strategies on VBench-Long.}
    \vspace{-1.5mm}
    \resizebox{\columnwidth}{!}{
        \setlength{\tabcolsep}{3pt}
        \renewcommand{\arraystretch}{1.1}
        \begin{tabular}{@{}l l | c c c c c c c c | c c@{}}
            \toprule
            \multirow{2}{*}{\makecell{KV Budget\\$(b_{\min}, b_{\max})$}} &
            \multirow{2}{*}{Mode} &
            \multirow{2}{*}{\makecell{Subject\\Consistency}} &
            \multirow{2}{*}{\makecell{Background\\Consistency}} &
            \multirow{2}{*}{\makecell{Motion\\Smoothness}} &
            \multirow{2}{*}{\makecell{Dynamic\\Degree}} &
            \multirow{2}{*}{\makecell{Aesthetic\\Quality}} &
            \multirow{2}{*}{\makecell{Imaging\\Quality}} &
            \multirow{2}{*}{\makecell{Overall\\Consistency}} &
            \multirow{2}{*}{\makecell{Clip\\Score}} &
            \multirow{2}{*}{\makecell{Visual\\Quality}} &
            \multirow{2}{*}{\makecell{Text\\Alignment}} \\
            & & & & & & & & & & & \\
            \midrule
            Self Forcing & --
            & 96.61 & 96.17 & 98.27 & 41.55 & 58.83 & 68.04 & 23.93 & 32.12 & 76.58 & 28.02 \\
            \midrule
            (4, 4)   & Normal
            & 91.93 & 93.88 & 97.55 & 42.78 & 60.15 & 68.24 & 25.86 & 33.78 & 75.76 & 29.82 \\
            (8, 8)   & Normal
            & 96.93 & 95.94 & 98.23 & 53.28 & 59.86 & 68.88 & 24.25 & 32.80 & 78.85 & 28.53 \\
            (12, 12) & Normal
            & 97.13 & 96.10 & 98.33 & 52.03 & 60.37 & 69.24 & 24.22 & 32.89 & 78.86 & 28.55 \\
            (4, 12)  & Normal
            & 95.84 & 95.49 & 98.04 & 61.39 & 60.59 & 68.62 & 24.53 & 32.97 & 80.00 & 28.75 \\
            (4, 12)  & Reverse
            & 97.00 & 95.90 & 98.28 & 52.60 & 60.11 & 68.85 & 24.33 & 32.87 & 78.79 & 28.60 \\
            (4, 12)  & Random
            & 96.82 & 95.86 & 98.26 & 52.93 & 60.40 & 69.05 & 23.92 & 32.72 & 78.89 & 28.32 \\
            \bottomrule
        \end{tabular}
    }
    \label{tab:budget_allocation_strategy_ablation}
    \vspace{-4mm}
\end{table}

\subsection{User Study}
\begin{figure}[htbp]
    \vspace{-2mm}
    \centering
    \includegraphics[width=0.72\linewidth]
    {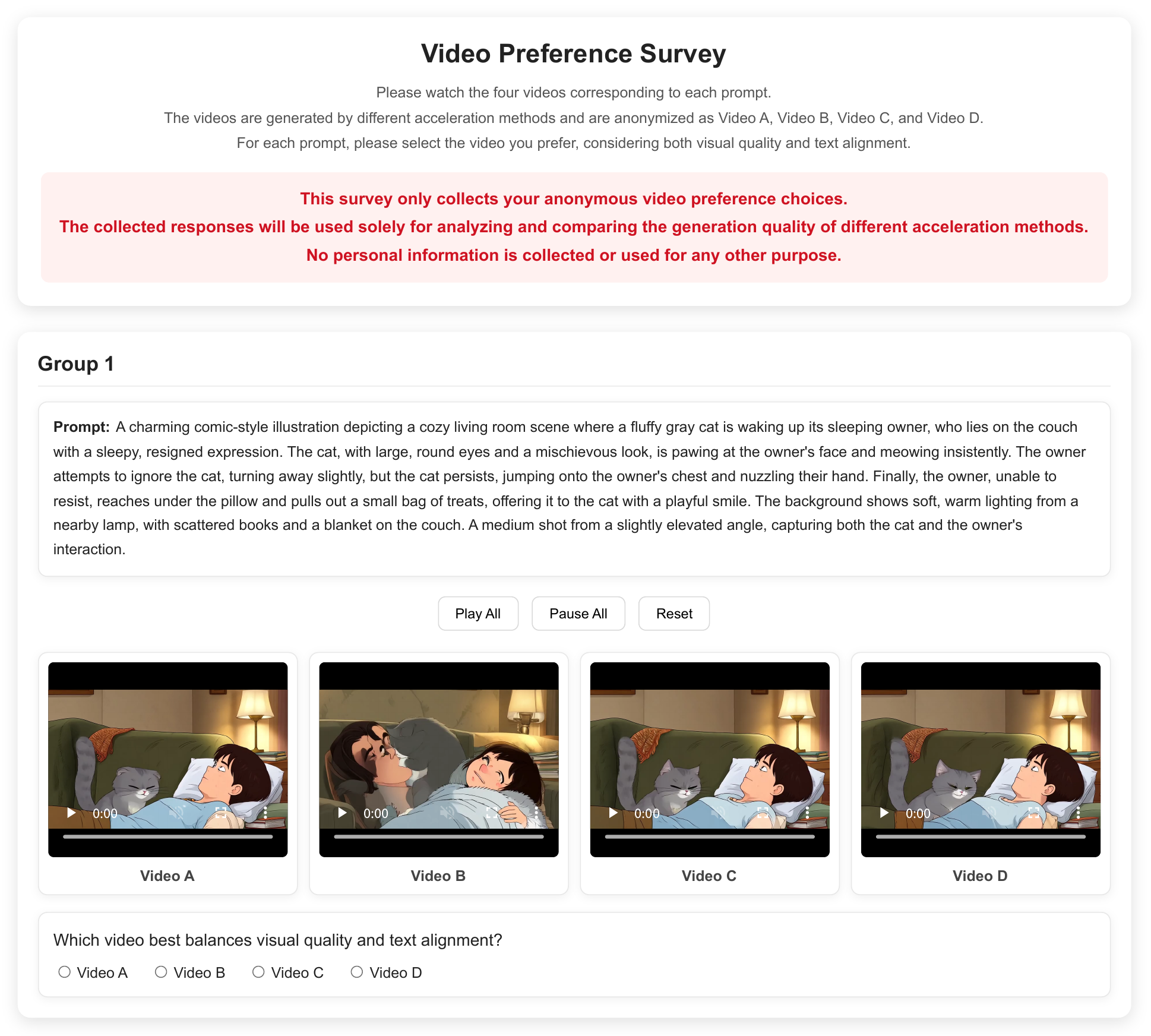}
    \vspace{-2mm}
    \caption{User study interface for comparing videos generated by different acceleration methods.}
    \label{fig:user_study}
    \vspace{-2mm}
\end{figure}

We conduct a user study to evaluate the perceptual quality of different acceleration methods under the chunk-wise Self Forcing paradigm. As shown in Fig.~\ref{fig:user_study}, each group presents four anonymized videos generated from the same prompt by MonarchRT, TaylorSeer, Dummy Forcing, and our method. Sixteen participants evaluate 10 groups, resulting in 160 preference votes. For each group, participants select the video that best balances visual quality and text alignment.

As reported in Tab.~\ref{tab:user-study}, our method achieves the highest preference ratio of 57.50\%, outperforming MonarchRT, TaylorSeer, and Dummy Forcing. This indicates that our method is also favored in subjective evaluation.
\begin{table}[H]
    \vspace{-1.5mm}
    \centering
    \caption{User study of different acceleration methods.}
    \vspace{-1.5mm}
    \resizebox{0.4\columnwidth}{!}{
        \setlength{\tabcolsep}{6pt}
        \renewcommand{\arraystretch}{1.1}
        \begin{tabular}{l c}
            \toprule
            \textbf{Methods} & \textbf{User Preference (\%)} \\
            \midrule
            MonarchRT & 18.12 \\
            TaylorSeer & 20.00 \\
            Dummy Forcing & 4.38 \\
            \rowcolor{gray!15}
            \textbf{Ours} & \textbf{57.50} \\
            \bottomrule
        \end{tabular}
    }
    \label{tab:user-study}
    \vspace{-4mm}
\end{table}

\subsection{Additional Qualitative Results}
\label{app:qualitative}
\begin{figure}[htbp]
    \vspace{-2mm}
    \centering
    \includegraphics[width=\linewidth]
    {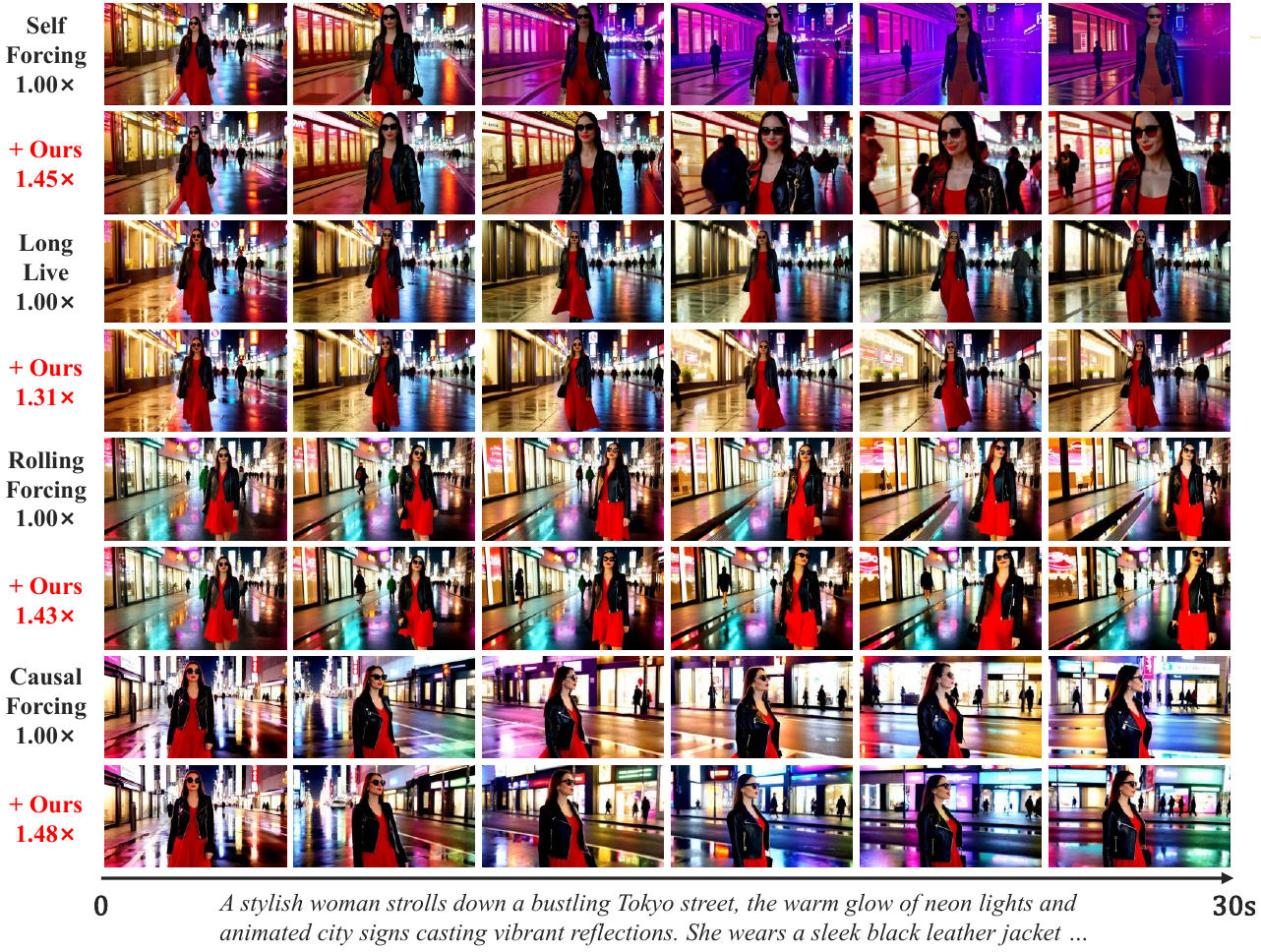}
    \vspace{-6mm}
    \caption{Qualitative examples comparing long-horizon consistency across different paradigms.}
    \label{fig:quality_comparison_0}
    \vspace{-2mm}
\end{figure}

\begin{figure}[htbp]
    \vspace{-2mm}
    \centering
    \includegraphics[width=\linewidth]
    {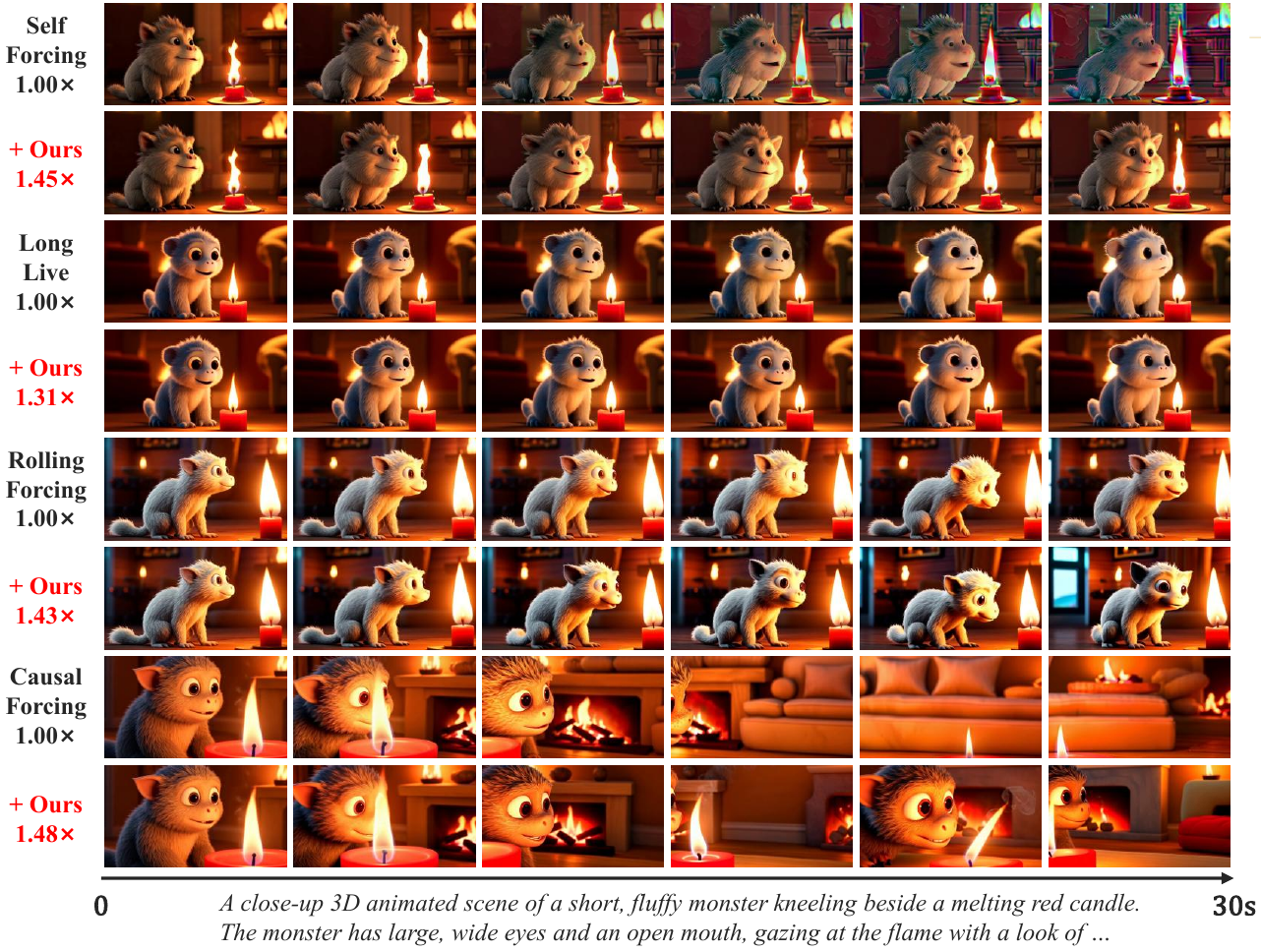}
    \vspace{-6mm}
    \caption{Qualitative examples comparing long-horizon consistency across different paradigms.}
    \label{fig:quality_comparison_1}
    \vspace{-2mm}
\end{figure}

\begin{figure}[htbp]
    \vspace{-2mm}
    \centering
    \includegraphics[width=\linewidth]
    {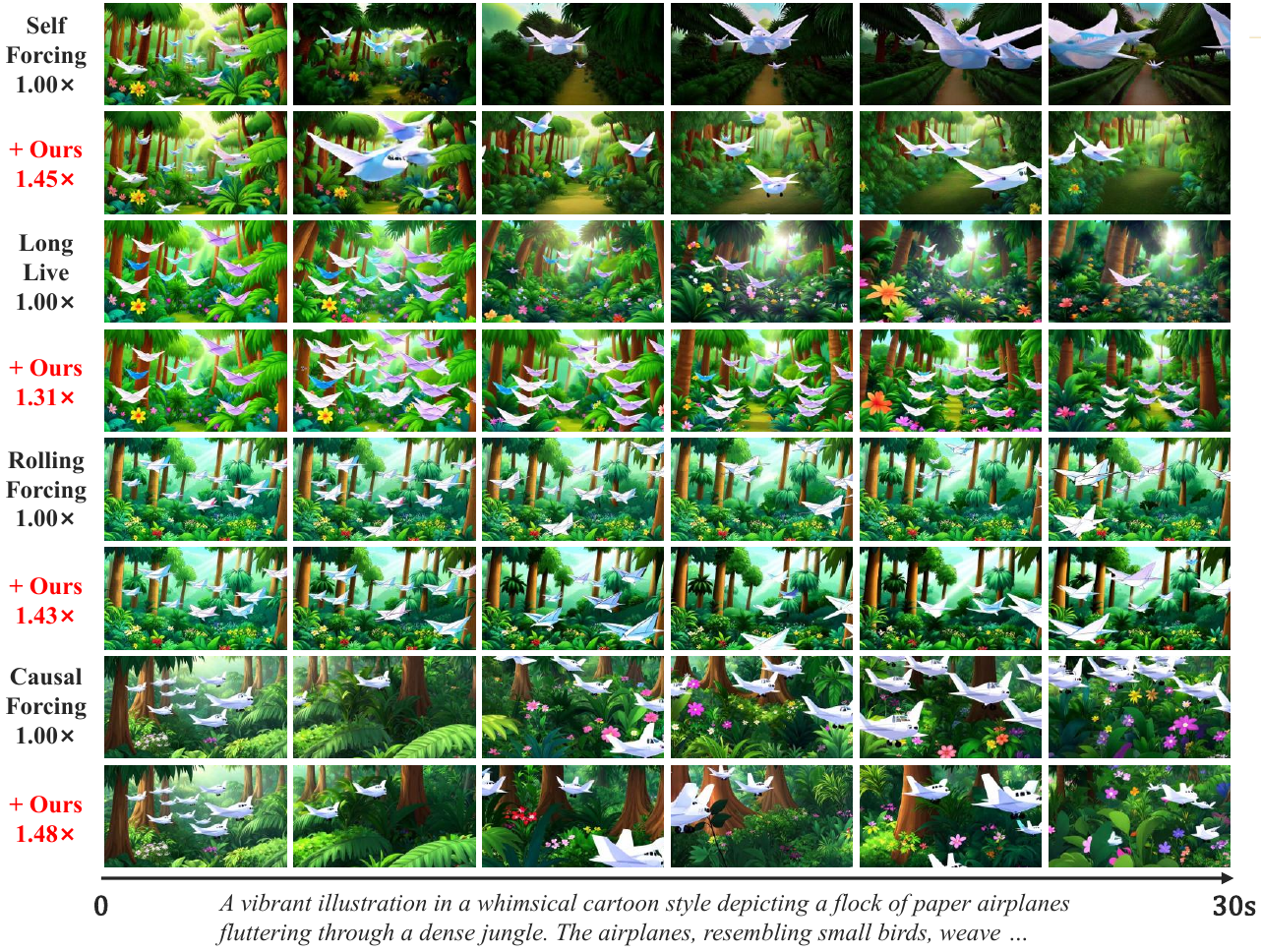}
    \vspace{-6mm}
    \caption{Qualitative examples comparing long-horizon consistency across different paradigms.}
    \label{fig:quality_comparison_2}
    \vspace{-2mm}
\end{figure}

\begin{figure}[htbp]
    \vspace{-2mm}
    \centering
    \includegraphics[width=\linewidth]
    {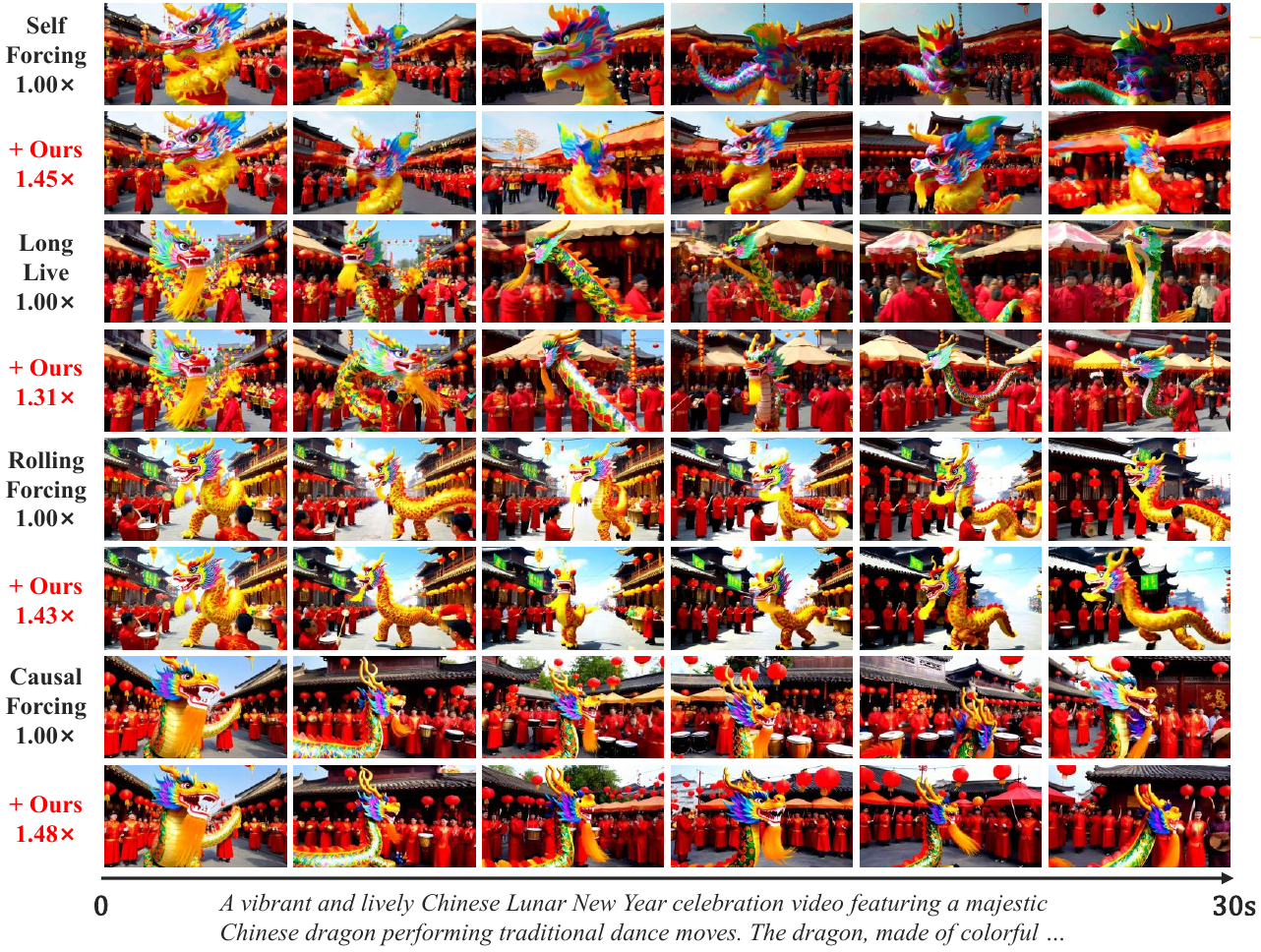}
    \vspace{-6mm}
    \caption{Qualitative examples comparing long-horizon consistency across different paradigms.}
    \label{fig:quality_comparison_3}
    \vspace{-2mm}
\end{figure}


\begin{figure}[htbp]
    \vspace{-2mm}
    \centering
    \includegraphics[width=\linewidth]
    {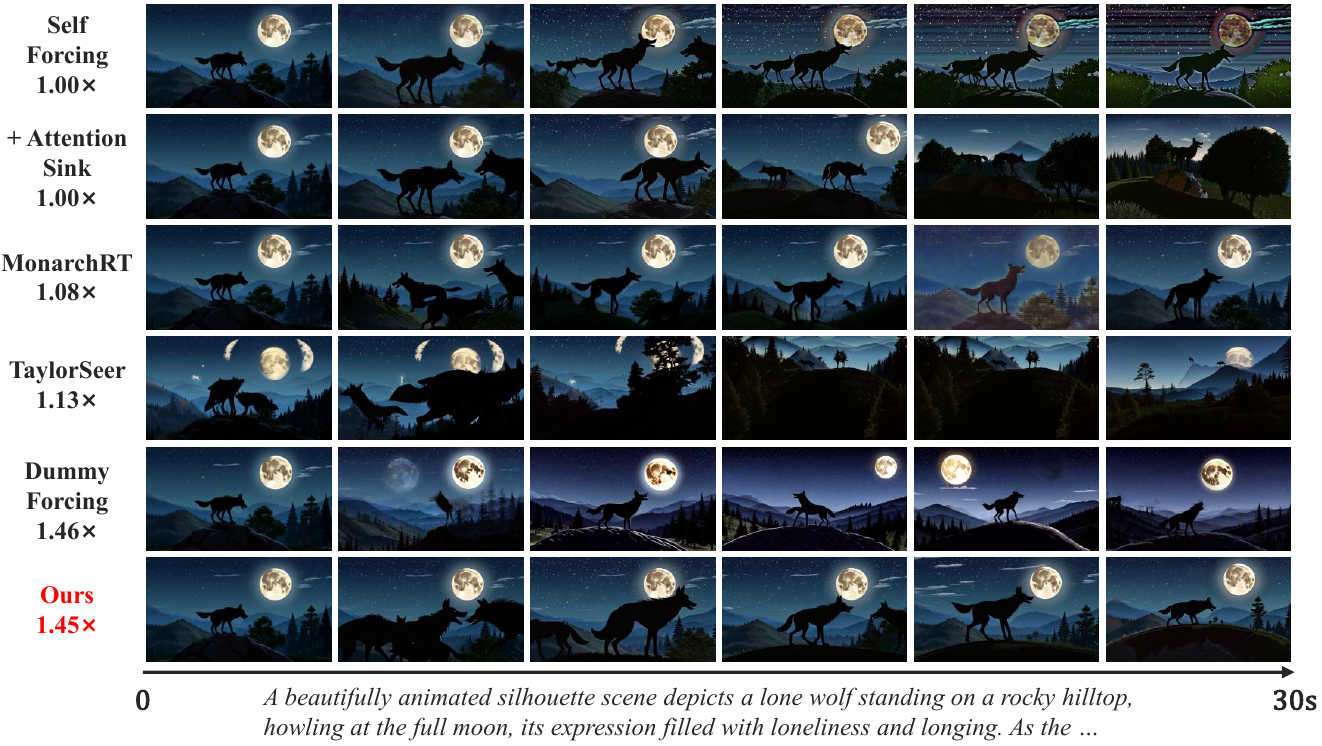}
    \vspace{-6mm}
    \caption{Qualitative examples comparing quality preservation across acceleration methods.}
    \label{fig:efficiency_comparison_1}
    \vspace{-2mm}
\end{figure}

\begin{figure}[htbp]
    \vspace{-2mm}
    \centering
    \includegraphics[width=\linewidth]
    {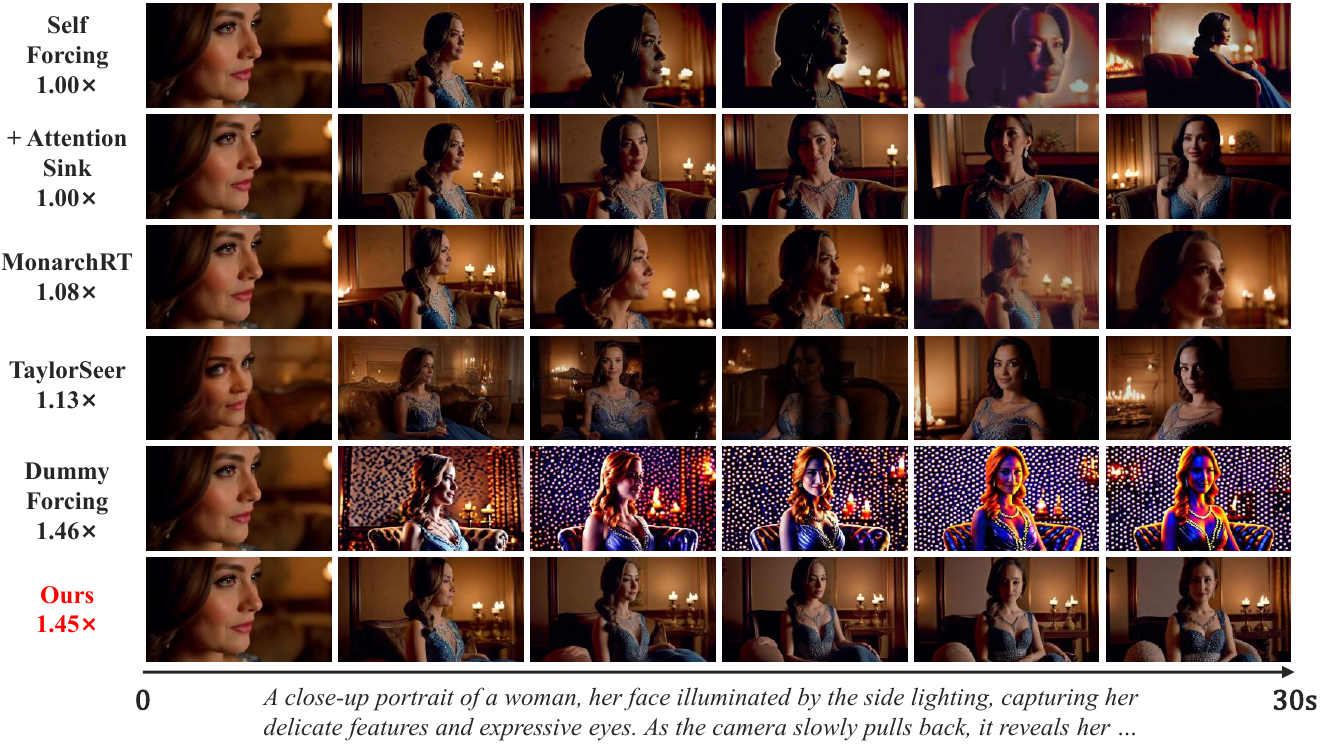}
    \vspace{-6mm}
    \caption{Qualitative examples comparing quality preservation across acceleration methods.}
    \label{fig:efficiency_comparison_2}
    \vspace{-2mm}
\end{figure}

\begin{figure}[htbp]
    \vspace{-2mm}
    \centering
    \includegraphics[width=\linewidth]
    {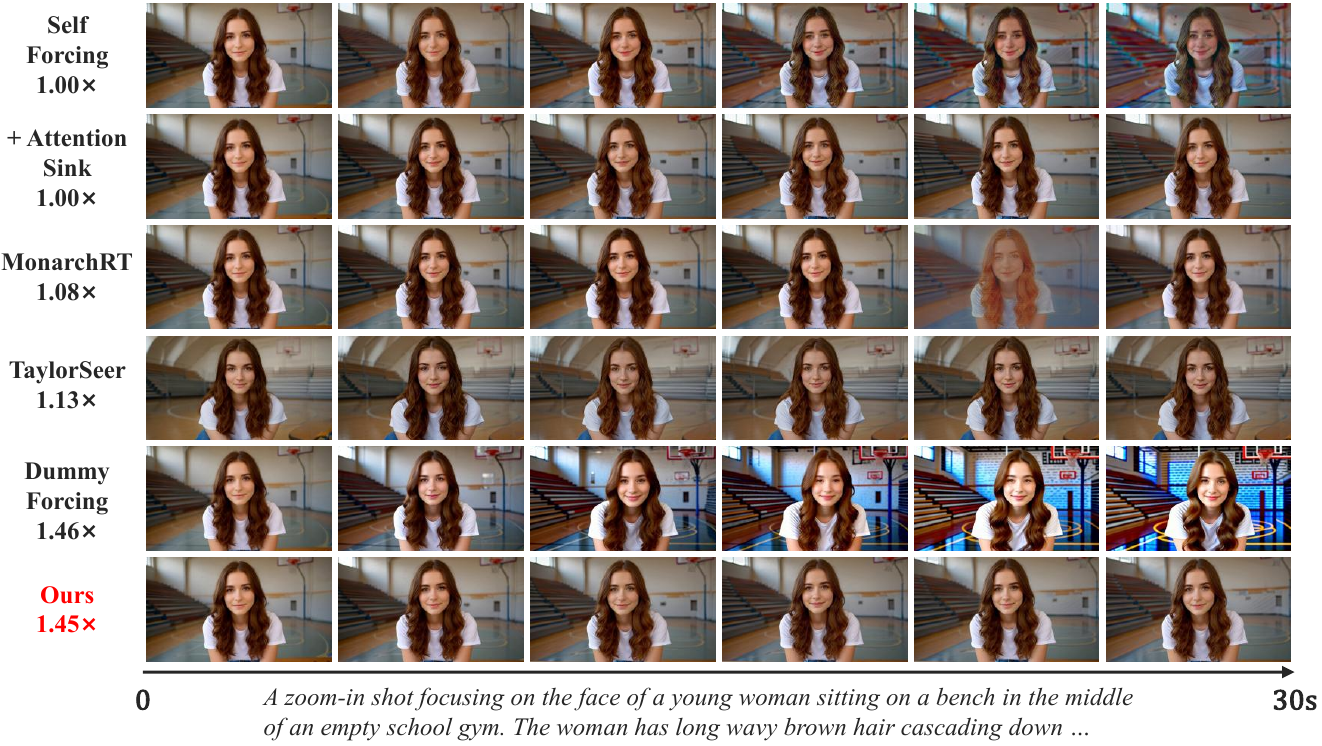}
    \vspace{-6mm}
    \caption{Qualitative examples comparing quality preservation across acceleration methods.}
    \label{fig:efficiency_comparison_3}
    \vspace{-2mm}
\end{figure}

\begin{figure}[htbp]
    \vspace{-2mm}
    \centering
    \includegraphics[width=\linewidth]
    {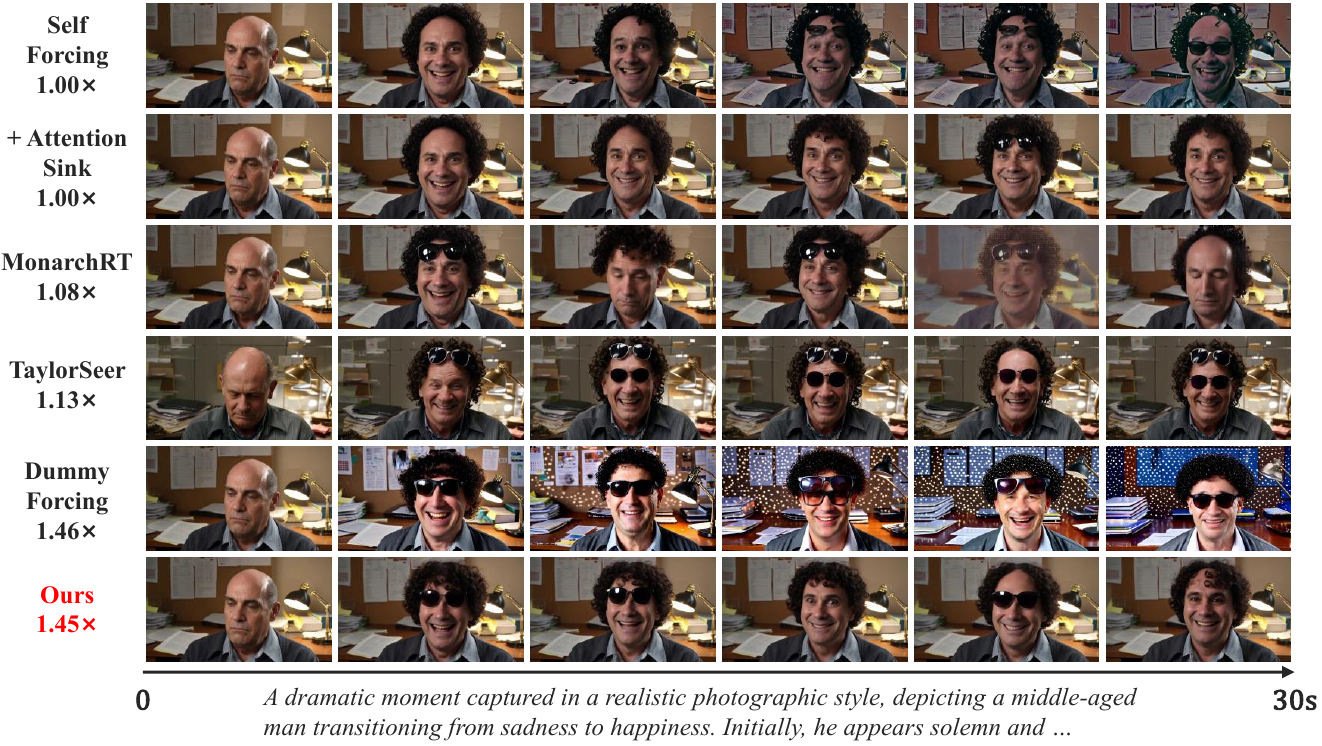}
    \vspace{-6mm}
    \caption{Qualitative examples comparing quality preservation across acceleration methods.}
    \label{fig:efficiency_comparison_4}
    \vspace{-2mm}
\end{figure}

\end{document}